\documentclass[runningheads]{llncs}

\usepackage{eccv}

\usepackage{eccvabbrv}

\usepackage{graphicx}
\usepackage{booktabs}
\usepackage{multirow}
\usepackage{wrapfig}

\usepackage[accsupp]{axessibility}  % Improves PDF readability for those with disabilities.

\usepackage{algorithm}
\usepackage{algorithmic}

\usepackage{hyperref}

\usepackage{orcidlink}

\newcommand{\datasetS}{\mathcal{S}}
\newcommand{\datasetD}{\mathcal{D}}
\begin{document}

\title{Leveraging Hierarchical Feature Sharing for Efficient Dataset Condensation}


\author{Haizhong Zheng\inst{1} \and
Jiachen Sun\inst{1}
\and Shutong Wu\inst{2}
\and Bhavya Kailkhura\inst{3}
\and
\\ Z. Morley Mao\inst{1}
\and Chaowei Xiao\inst{2}
\and Atul Prakash\inst{1}
}

\authorrunning{H. Zheng et al.}

\institute{University of Michigan, Ann Arbor
\and University of Wisconsin, Madison
\and Lawrence Livermore National Laboratory\\
% \email{\{a,lncs\}@uni-heidelberg.de}
}

\maketitle

% \vspace{-0.2cm}
\begin{abstract}
% \ap{
% One sentence on importance of datasate condensation
% Don't mention data parametrization. Instead, mention the challenge faced by the best schemes today. Then say what our intuition or technique is.
% Then, point out a key advantage -- quantifying if possible.
% Also, I am not a fan of "HMN" term. I wish it was renamed.
% The reason is that it is not a full tree. Instead, the main intuition over prior art is dataset-level and class-level sharing, whereas data parametrization does instance-level sharing (if I understand things right).  So, maybe that should be bought up as the main concept. E.g., GPS (Group Property Sharing) could be a better title.
% }

Given a real-world dataset, data condensation (DC) aims to synthesize a small synthetic dataset that captures the knowledge of a natural dataset while being usable for training models with comparable accuracy. 
Recent works propose to enhance DC with {\em data parameterization}, which condenses data into very compact parameterized data containers instead of images. 
The intuition behind data parameterization is to 
encode \emph{shared features} of images to avoid additional storage costs.
In this paper, we recognize that images share common features in a hierarchical way due to the inherent hierarchical structure of the classification system, which is overlooked by current data parameterization methods.
To better align DC with this hierarchical nature and encourage more efficient information sharing inside data containers, we propose a novel data parameterization architecture, \emph{Hierarchical Memory Network (HMN)}.
HMN stores condensed data in a three-tier structure, representing the dataset-level, class-level, and instance-level features.
Another helpful property of the hierarchical architecture is that HMN naturally ensures good independence among images despite achieving information sharing.
This enables instance-level pruning for HMN to reduce redundant information, thereby further minimizing redundancy and enhancing performance.
We evaluate HMN on five public datasets and show that our proposed method outperforms all baselines.

\end{abstract}

% \vspace{-0.8cm}
\section{Introduction}
\vspace{-0.1cm}

%(1) Define data condensation problem, introducing the term "memory budget". (2)  Currently, there are two orthogonal techniques. Say something about pros and cons and how far they are able to condense data. Synthetic data set versus a coreset selection (natural subset). For CIFAR-10, for 5 image per class, using synthetic data set 

%Data condensation methods, which use synthetic datasets, significantly outperform in terms of accuracy  another class of methods called coreset selection that select a subset of the dataset $D$ 

%Data distillation, also known as data condensation, is a technique introduced by Wang et al.~\cite{wang2018dataset} with the goal of this method is to create a smaller, synthesized dataset that effectively represents the original, larger training dataset. 

% Wang et al.~\cite{wang2018dataset} proposed {\em data condensation} (also called data distillation) as a technique to improve deep learning data efficiency.

Introduced by Wang et al.~\cite{wang2018dataset}, given a training dataset $\datasetD$, the aim of {\em data condensation}~(DC), also known as data distillation, is to synthesize a much smaller {\em synthetic dataset} $\datasetS$ such that $\datasetS$ can be used to train models that are comparable in test data performance to those trained on $\datasetD$.
Given increasing sizes of datasets, DC has emerged as an important goal for compute- and storage-efficient deep learning~\cite{bartoldson2023compute, zheng2020efficient, du2022minimizing, nguyen2021dataset, nguyendataset, shin2023loss, cui2023scaling}. 
Researchers have shown that DC can provide significant efficiencies in diverse applications such as continual learning~\cite{rosasco2022distilled, sangermano2022sample}, network architecture search~\cite{zhao2023dm}, and federated learning\cite{song2022federated, xiong2022feddm}.

To improve the effectiveness of DC methods, Kim et al.~\cite{kim2022dataset} propose \textbf{data parameterization}. 
Instead of condensing data into images, data parameterization condenses data into parameterized data containers.
Such a data container is a parameterized function $f_\theta$ such that $f_\theta(\cdot)$ generates the synthetic dataset $\datasetS$. The goal is that storing $f_\theta$, represented by its parameters $\theta$, is much more compact than storing $\datasetS$.
The intuition behind data parameterization methods is to 
encode \emph{shared features} among images together into a {\em data container} to make the data condensation more effective~\cite{dengremember, liudataset}.

Recognizing this shared feature insight, it's important to delve deeper into the inherent structure of these shared features in datasets.
We notice that images share common features in a hierarchical way due to the inherent hierarchical structure of the classification system.
Even if images differ in content, they can still share features at different hierarchical levels.
For example, two images of cats can share common features specific to the cat class, but an image of a cat and another of a dog may still have shared features of the broader animal class.

\begin{wrapfigure}{r}{0.48\textwidth}
    \vspace{-0.15cm}
    \centering
    \includegraphics[width=\linewidth]{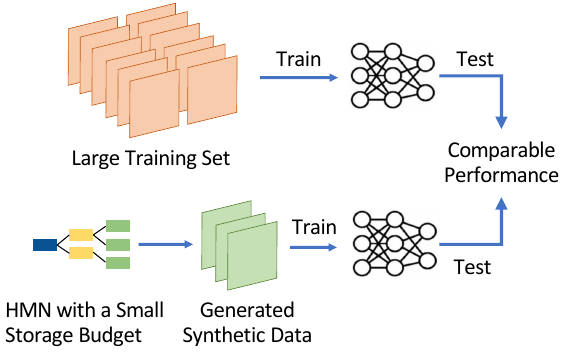}
    \caption{ Illustration of data condensation with HMN.
    Like other data parameterization methods, HMN is a data container using a small storage budget and can generate images for training.}
    \label{fig:HMN-concept}
    \vspace{-0.5cm}
\end{wrapfigure}
\begin{figure}[t]
    \centering
    % \vspace{-0.2cm}
    \includegraphics[width=\linewidth]{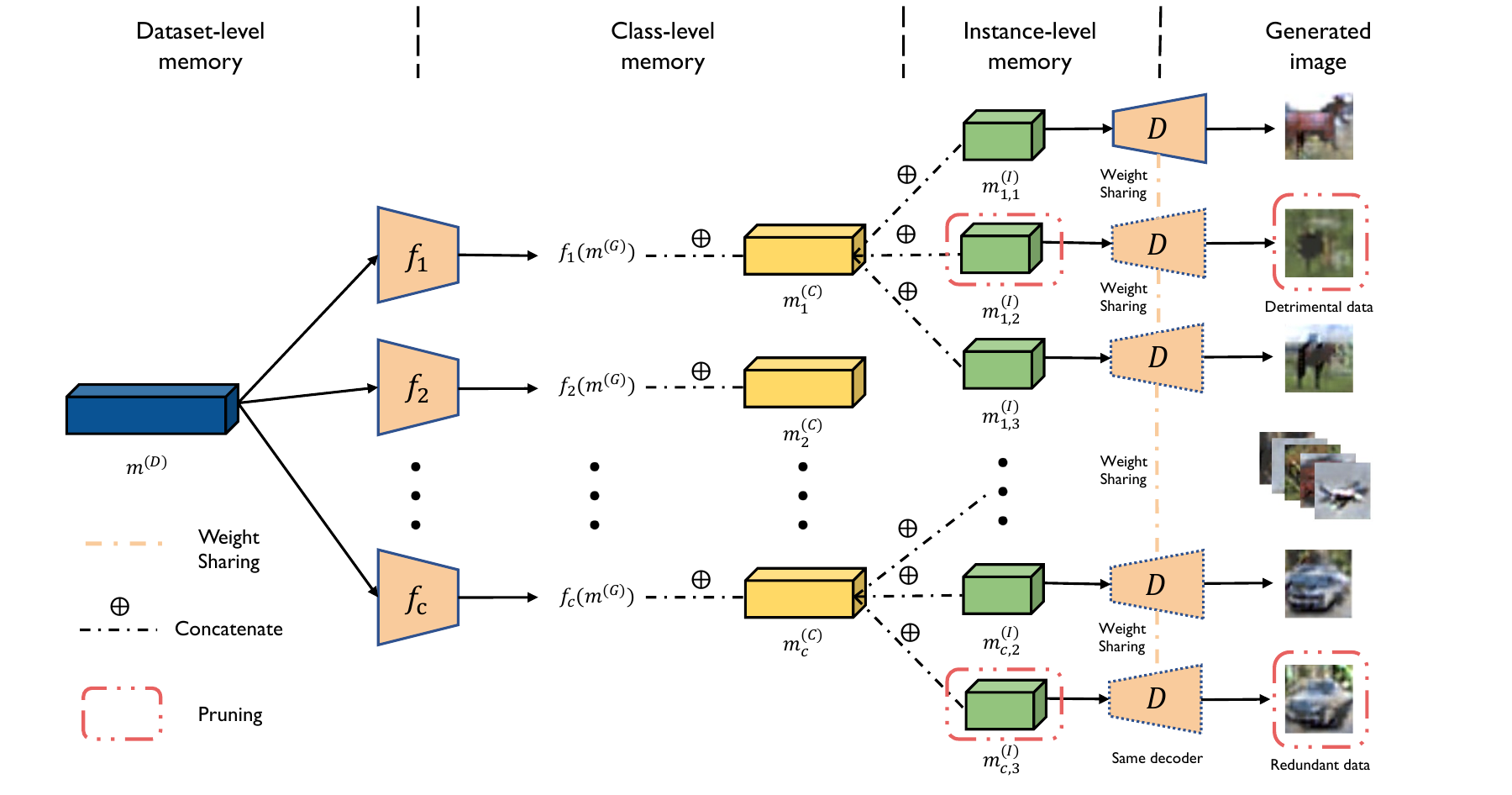}    
    \caption{Illustration of Hierarchical Memory Network and pruning. HMN consists of three tiers of memories (which are learnable parameters). $f_i$ is the feature extractor for each class. $D$ is a single shared decoder to translate a concatenated memory to a synthetic image, though it is applied on a per-image basis, as shown. When we identify redundant or detrimental images, the corresponding instance-level memories are pruned, as indicated by red boxes, saving storage budget. }
    \label{fig:HMN}
   % \vspace{-1.5cm}
\end{figure}

However, current data parameterization methods that adopt factorization to share features among images overlook this hierarchical nature of shared features in datasets.
In this paper, to better align with this hierarchical nature and encourage more efficient information sharing inside data containers, we propose a novel data parameterization architecture, \emph{Hierarchical Memory Network (HMN)}. 
Figure~\ref{fig:HMN-concept} illustrates how HMN is used for data condensation. An HMN can be used to efficiently generate a synthetic dataset, which can then be used to train a model that is designed to be close in performance to a model that is trained on a larger dataset. Zooming in on the HMN, as illustrated in Figure~\ref{fig:HMN}, an HMN comprises a three-tier memory structure: \emph{dataset-level memory}, \emph{class-level memory}, and \emph{instance-level memory}.
Examples generated by HMNs share information via common dataset-level and class-level memories. 
Another helpful property of the hierarchical architecture is that HMN naturally ensures good independence among images.
We find that condensed datasets contain redundant data, 
indicating room for further improvement in data condensation by pruning redundant data. 
In Section~\ref{ssec:pruning}, we show that, compared to other data containers, HMNs are easier to prune. We propose an algorithm to prune HMNs to further reduce the redundancy in HMNs.

We evaluate our proposed methods on four public datasets (SVHN, CIFAR10, CIFAR100, and Tiny-ImageNet) and compare HMN with the other nine baselines.
The evaluation results show that, even when trained with a low GPU memory consumption batch-based loss, HMN still outperforms all baselines, including those using high GPU memory trajectory-based losses.
For a fair comparison, we also compare HMN with other data parameterization baselines under the same loss.
We find that HMN outperforms these baselines by a larger margin.
For instance, HMN outperforms at least 3.7\%/5.9\%/2.4\% than other data parameterization methods within 1/10/50 IPC~(Image Per Class)\footnote{IPC measures the equivalence of a tensor storage budget in terms of the number of images. For example,  1 IPC for CIFAR10 stands for:  Pixels of an image * IPC * class = 3 * 32 * 32 * 1 * 10 = 30720 tensors. The same metric is also used in SOTA works~\cite{liudataset, dengremember}.}
storage budgets when trained with the same loss on CIFAR10, respectively.
Additionally, we also apply HMN to continual learning tasks. The evaluation results show that HMNs effectively improve the performance on continual learning.

% \vspace{-0.25cm}
\section{Background and Related Work}
\label{sec:related-work}
% \vspace{-0.2cm}

\noindent{\bf Problem Definition:} As described in Wang et al.~\cite{wang2018dataset}, given an original training dataset $\datasetD$, DC methods aim to generate a synthetic dataset $\datasetS$, which uses a much smaller storage than $\datasetD$, i.e., $|\datasetS| << |\datasetD|$, while achieving a comparable  performance (accuracy) of models trained on $\datasetS$ to that on $\datasetD$.

% Given a dataset $\mathcal{T}$, data condensation {\em synthesizes} a much smaller dataset $\mathcal{S}$ that trains the model to achieve comparable performance to the one trained with $\mathcal{T}$. 

\noindent{\bf Training Loss Functions for Optimizing Synthetic Datasets:} A common element of existing approaches to data condensation is to choose a training loss function that helps optimize the performance gap between synthetic and real dataset. Two main types of training loss are used to optimize synthetic datasets:  
1) \emph{batch-based loss}~ \cite{zhao2023dm, zhao2021dataset2, zhao2021dataset}, and 
2) \emph{training trajectory-based loss}~\cite{wang2018dataset, cazenavette2022dataset}. 

Zhao et al.~\cite{zhao2021dataset} proposed a batch-based loss method called {\em gradient matching} that aims to minimize the distance between the gradients of a batch of synthetic data and original data.  
Another batch-based loss method is distribution matching~\cite{zhao2023dm} that aims to minimize the distance between the embeddings of a batch of synthetic dataset $\datasetS$ and original dataset $\datasetD$. 
% IDC~\cite{kim2022dataset} and IDM~\cite{zhao2023improved}
% state-of-the-art batch-based loss method techniques.

In contrast to batch-based loss methods,  
trajectory loss requires training the model on the synthetic dataset for multiple iterations while monitoring how the synthetic dataset updates the model parameters across iterations.
MTT~\cite{cazenavette2022dataset} employs the distance between model parameters of models trained on the synthetic dataset $\datasetS$ and those trained on the original dataset $\datasetD$ as the loss metric.
Trajectory-based losses generally tend to provide better empirical performance than batch-based losses, but have considerably larger GPU memory consumption~\cite{cazenavette2022dataset, cuidc}. 
{\em Our work demonstrates that by employing the HMN architecture as a data container, a batch-based loss can achieve comparable and even better performance than current data container methods that are based on memory-intensive trajectory-based loss.}

\noindent\textbf{Data Parameterization for Data Condensation.}
Given a storage budget, \emph{data parameterization}~\cite{dengremember, liudataset, kim2022dataset} has been recently proposed to further improve data condensation over just optimizing a synthetic dataset $\datasetS$. 
Instead of condensing data into image space, the key idea of data parameterization is to condense data into compact free-parameter data containers that can generate training images.
Data parameterization allows different generated training images to share the same weights in the data containers, which improves storage efficiency.
IDC~\cite{kim2022dataset} proposes to downsample images to improve storage efficiency.
HaBa~\cite{liudataset} and LinBa~\cite{dengremember} concurrently introduce factorization-based methods to improve data condensation by sharing common information among images.

\noindent{\bf Other Methods:} Some recent work~\cite{cazenavette2023generalizing, zhao2022synthesizing} explores generating condensed datasets with generative priors~\cite{brock2017neural, chaiusing, chai2021ensembling}. For example, instead of synthesizing the condensed dataset from scratch, GLaD~\cite{cazenavette2023generalizing} assumes the existence of a well-trained generative model. We do not assume the availability of such a generative model and thus this line of work is beyond the scope of this paper, though it is worthy of future investigation.

\noindent{{\bf Coreset Selection} is another technique aimed at enhancing data efficiency~\cite{coleman2019selection, xiamoderate, li2023less, sener2017active, sorscher2022beyond}.  Rather than generating a synthetic dataset as in our work, coreset selection identifies a representative subset from the original dataset. Unfortunately, coresets do not tend to give as much data condensation as synthetic datasets and thus we focus on generating synthetic datasets in this work. Nevertheless, coreset selection methods such as the area under the margin (AUM)~\cite{pleiss2020identifying} that measure the data importance by accumulating output margin across training epochs are useful as an initial step in synthetic data condensation as they can be used to select more representative portion of the dataset $D$ to initialize condensed data synthesis~\cite{cuidc, liu2023dream}.

\vspace{-0.3cm}
\section{Methodology}
\vspace{-0.2cm}

In this section, we present technical details on the proposed data condensation approach. 
In Section~\ref{ssec:hmn}, we present the architecture design of our novel data container for condensation, Hierarchical Memory Network (HMN), to better align with the hierarchical nature of common feature sharing in datasets. 
In Section~\ref{ssec:pruning}, we study data redundancy of datasets generated by data parameterization methods and show that techniques inspired by coreset methods can be used to prune an HMN data container to make them more storage-efficient. 

% We introduce the HMN architecture next, highlighting its efficiency in delivering high model performance on a constrained storage budget and lower GPU memory costs compared to current leading data parameterization methods. Performance details are discussed in Section~\ref{sec:experiments}.
% We next describe the HMN architecture (Section~\ref{ssec:hmn}).
% that provides compelling advantages in that, given a storage budget, it is able to achieve high model performance at lower GPU memory costs than SOTA data parametrization approaches (performance results are later in Section~\ref{sec:experiments}). 
% We then study the data redundancy of datasets generated by data parameterization methods and show that techniques inspired by coreset methods can be used to prune an HMN data container to make them highly storage-efficient (Section~\ref{ssec:pruning}). 

% \textbf{Overview.} Our proposed data condensation approach consists of two steps: 1) We first condense the dataset information into over-budget HMNs. 2) Then, we identify and prune redundant or detrimental information from the over-budget HMNs to get HMNs within a given storage budget.

% \vspace{-0.2cm}
\subsection{Hierarchical Memory Network (HMN)}
\label{ssec:hmn}

A Hierarchical Memory Network (HMN) is a parameterized data container such that given an image index $i$, it outputs a synthetic image $\datasetS_i$, where $1 \le i \le |\datasetS|$. Here $|\datasetS|$ is still a small fraction of $|\datasetD|$. Naturally, it can also be used to generate the entire synthetic dataset $\datasetS$.

HMN is inspired by earlier work on data parameterization in that it takes advantage of shared features within the dataset $\datasetD$. However, HMN goes further in that it exploits shared common features from a hierarchical perspective. 
For instance, two images of cats can share common features specific to the cat class, but an image of a cat and another of a dog may still have shared features of the broader animal class. 
Our key insight for HMN is that images from the same class can share class-level common features, and images from different classes can share dataset-level common features.
As shown in Figure~\ref{fig:HMN}, HMN is a three-tier hierarchical data container to store condensed information.  Each tier comprises one or more memory tensors, and memory tensors are learnable parameters. The three tiers are as follows:
\begin{enumerate}
    \item {\bf Dataset-level memory:} The first tier is a dataset-level memory, $m^{(\datasetD)}$, which stores the dataset-level information shared among images in the dataset.
\item {\bf Class-level memories:} The second tier, the class-level memory, $m^{(C)}_c$, where $c$ is the class index. The class-level memories store class-level shared features.  The number of class-level memories is equivalent to the number of classes in the dataset.
\item {\bf Instance-level memories:} The third tier stores the instance-level memory, $m^{(I)}_{c, i}$, where $c, i$ are the class index and instance index, respectively. 
The instance-level memories are designed to store unique information for each image. 
The number of instance-level memories determines the number of images the HMN generates for training.
\end{enumerate}

Besides the memory tensors, we also have feature extractors $f_i$ for each class and a uniform decoder $D$ to convert concatenated memory to images. 

{\bf Storage Budget for HMN:}  Following past work on data containers using parameterization~\cite{kim2022dataset,dengremember,liudataset},  \emph{ memory tensors, feature extractors,   and uniform decoder $D$  network weights count towards the storage budget.} 
The generated synthetic images, $\datasetS$, do not count towards the storage budget, since they can be generated as needed, like an efficient image lookup memory.

\textbf{Other Design Attempts.} In the preliminary stages of designing HMNs, we also considered applying feature extractors between $m^{(C)}_c$ and $m^{(I)}_{c, i}$, and attempted to use different decoders for each class to generate images.
However, introducing such additional networks did not empirically improve performance. 
In some cases, it even causes performance drops.
One explanation for these performance drops with an increased number of networks is overfitting: more parameters make a condensed dataset better fit the training data and specific model initialization but compromise the model's generalizability.
Consequently, we decided to only apply feature extractors on the dataset-level memory and use a uniform decoder to generate images.

To generate an image for class $c$, we first adopt features extractor $f_c$ to extract features from the dataset-level memory~\footnote{In some storage-limited settings, such as when storage budget is 1IPC, we utilize the identity function as $f_c$.}. 
This extraction is followed by a concatenation of these features with the class-level memory $m^{(C)}_c$ and instance-level memory $m^{(I)}_{c,i}$. The concatenated memory is then fed to a shared decoder $D$, which generates the image used for training.
Formally, the $i$th generated image, $x_{c,i}$, in the class $c$ is generated by the following formula:
\begin{equation}
    x_{c,i} = D([f_c(m^{(\datasetD)}) \oplus m^{(C)}_c \oplus m^{(I)}_{c,i}])
% \vspace{-0.1cm}
\end{equation}
% \ap{The above equation changed to use XOR operator between the terms, not semi-colons.}

We treat the size of memories and the number of instance-level memories as hyperparameters for architecture design.
We present design details in Appendix~C, including the shape of memories, the number of generated images per class, architectures of feature extractors and decoder. 
Given the same storage budget, HMNs generate more training images than other DC methods. 
However, more generated training images do not hurt training efficiency on the condensed datasets.
In Appendix~D.2, we show that training with HMNs achieves better accuracy given the same training time compared to SOTA methods.

% It is also important to note that, in contrast to classic generative models, HMN does not have any input, and the output of HMN is always fixed.

%\ap{I think this is not going to be clear here. Let's remove it and discuss in Ablation Study in more depth -- not clear what data condensation setting means here and why that is not applicable to factorization-based methods without more details.}}
% \ap{Consider deleting the following para. Or move it to ablation study. It doesn't fit well here and hard to understand without appropriate context from the ablation study.}
% We also observe that, in contrast to factorization-based data containers, the design of HMNs is more similar to classic neural networks and thus offers better flexibility in designing more adaptive data containers for different data condensation settings. For example, in section~\ref{ssec:ablation}, we investigate how the size of instance-level memory influences the data condensation performance.

\textbf{Training Loss.} HMN can be integrated with either trajectory-based loss or batch-based loss. In this paper, to avoid high GPU memory demands of trajectory-based loss, we use a batch-based loss measure of gradient matching~\cite{kim2022dataset} to condense information into HMNs.
Given the original dataset $\mathcal{T}$, the initial model parameter distribution $P_{\theta_0}$, a distance function $d$, and loss function $\mathcal{L}$, gradient matching aims to synthesize a dataset $\mathcal{S}$ by solving the following optimization:
\vspace{-0.1cm}
\begin{equation}
    \min_{\mathcal{S}} \mathbf{E}_{\theta_0 \sim P_{\theta_0}}[\sum_{t=0}^{T-1} d (\nabla_\theta \mathcal{L}(\theta_t, \mathcal{S}), \nabla_\theta \mathcal{L}(\theta_t, \mathcal{T}))],
\label{eq:gm-loss}
\end{equation}
%\vspace{-0.1cm}
where $\theta_t$ is learned from $\mathcal{T}$ based on $\theta_{t-1}$, and $t$ is the iteration number.
In our scenario, the condensed dataset $\mathcal{S}$ is generated by an HMN denoted as $H$.
In Section~\ref{ssec:performance-compaison}, our evaluation results show that our data condensation approach, even when employing a batch-based loss, achieves better performance than SOTA DC baselines, including those that utilize high-memory trajectory-based losses.

%\vspace{-0.2cm}
\subsection{Post-Condensation Pruning of an HMN}
\label{ssec:pruning}
\vspace{-0.1cm}

In this part, we first show that data redundancy exists in condensed synthetic datasets. Then, we propose a pruning algorithm on HMN to reduce such data redundancy. While such pruning, in theory, can be applied to other data containers, HMN is particularly suited to such pruning because of instance-level memory (see pruned boxes in Figure~\ref{fig:HMN}). 

% we are going to discuss the redundancy of condensed datasets, and discuss why pruning on SOTA data containers is hard and HMN naturally has good properties for pruning.

\vspace{-0.2cm}
\subsubsection{Data Redundancy in Condensed Datasets}
\label{sssec:method-redundancy}
\vspace{-0.1cm}

Real-world datasets are shown to contain many redundant data~\cite{zheng2022coverage, pleiss2020identifying, toneva2018empirical, zheng2024elfs}. 
Here, we show that such data redundancy also exists in condensed datasets. 
We use HaBa~\cite{liudataset} as an example.
We first measure the difficulty of training images generated by HaBa with the area under the margin (AUM)~\cite{pleiss2020identifying}, a metric measuring data difficulty/importance.
The margin for example $(\mathbf{x}, y)$ at training epoch $t$ is defined as:
\begin{equation}
% \vspace{-0.1cm}
M^{(t)}(\mathbf{x}, y) = z_y^{(t)}(\mathbf{x}) - \max_{i \neq y} z_i^{(t)}(\mathbf{x}),
\vspace{-0.1cm}
\end{equation}

where $z_i^{(t)}(\mathbf{x})$ is the prediction likelihood for class $i$ at training epoch $t$.
AUM is the accumulated margin across all training epochs:
\vspace{-0.1cm}
\begin{equation}
    \mathbf{AUM}(\mathbf{x}, y) = \frac{1}{T} \sum_{t=1}^T M^{(t)}(\mathbf{x}, y).
\label{eq:aum}
%\vspace{-0.7cm}
\end{equation}

\begin{table}
    \centering
    % \scriptsize
    \vspace{-0.8cm}
    \setlength{\tabcolsep}{3.2pt}
    \caption{Coreset selection on the training dataset generated by HaBa on CIFAR10 10 IPC. The data with high AUM is pruned first.}
    \begin{tabular}{c|ccccc}
        \hline
        Pruning Rate &  0 & 10\% & 20\% & 30\% & 40\% \\
        \hline
        Accuracy (\%) & 69.5 & 69.5 & 68.9 & 67.6 & 65.6 \\
        \hline
    \end{tabular}
    \vspace{-0.3cm}
    \label{tab:haba-pruning}
\end{table}

\begin{wrapfigure}{r}{0.45\textwidth}
    \vspace{-0.7cm}
    \centering
    \includegraphics[width=\linewidth]{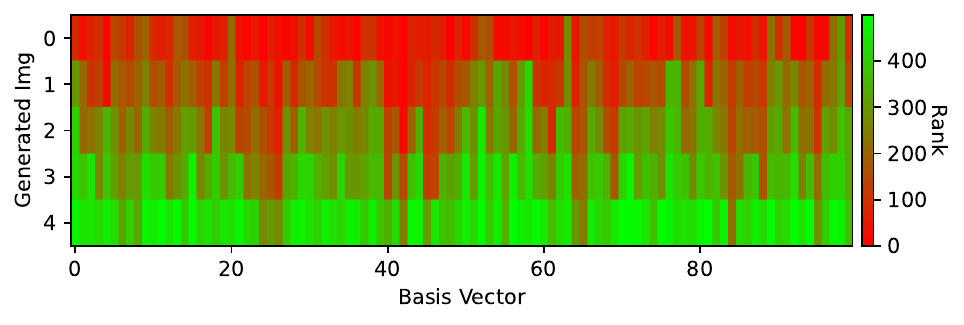}
    \caption{Rank distribution for different basis vectors in HaBa for CIFAR10 10 IPC. Each column in this figure represents the difficulty rank of images generated using the same basis vector. The color stands for the difficulty rank among all generated images. Green denotes easy-to-learn images, while red indicates hard-to-learn images. }
    % The difficulty of images generated by the same basis vector can be very different. Thus, simply eliminating a basis vector does not guarantee selective pruning of only the desired images.}
    \label{fig:haba-basis-difficulty}
    % \vspace{-0.2cm}
\end{wrapfigure}

A low AUM value indicates that examples are hard to learn. Those examples with lower AUM value are harder to learn, thus are thought to provide more information for training and are more important\cite{toneva2018empirical, pleiss2020identifying, zheng2022coverage}.
Then, as suggested in SOTA coreset selection work~\cite{toneva2018empirical}, we prune out the data with smaller importance (high AUM). 
The results of coreset selection on the dataset generated by HaBa for CIFAR10 10 IPC are presented in Table~\ref{tab:haba-pruning}. 
We find that pruning up to 10\% of the training examples does not hurt accuracy. 
This suggests that these 10\% examples are redundant and can be pruned to save the storage budget. 

Pruning on generated datasets is straightforward, but pruning relevant weights in data containers proposed in prior work can be challenging because of dependency among weights in data containers. 
% (We include an example in Appendix~\ref{}).
Unlike prior containers, pruning an HMN is straightforward since each generated image has its own instance-level memory, which allows us to prune redundant generated images by pruning corresponding instance-level memories (as illustrated by red boxes in Figure~\ref{fig:HMN}).
%SOTA well-performed data parameterization methods, like LinBa and HaBa, use factorization-based methods to generate images. 
%Factorization-based methods use different combinations between basis vectors and decoders to share information,
%but this also creates interdependence among images, making pruning single images challenging.

A potential solution for pruning factorization-based data containers is to prune basis vectors in the data containers (each basis vector is used to generate multiple training images).
However, we show that directly pruning these basis vectors can lead to removing important data.
In Figure~\ref{fig:haba-basis-difficulty}, we plot the importance rank distribution for training data generated by each basis vector.
We observe that the difficulty of images generated by the same basis vector can differ greatly. Thus, simply pruning basis vectors does not guarantee selective pruning of only desired images.

\vspace{-0.5cm}
\subsubsection{Over-budget Condensation and Post-Condensation Pruning}
\label{sssec:method-pruning}
% \vspace{-0.15cm}
To condense datasets with specific storage budgets and take advantage of the pruning property of HMN to further enhance data condensation, we propose to first condense data into over-budget HMNs, which exceed the storage budget by $p\%$~($p$ is a hyperparameter). 
Then, we prune these HMNs to fit the allocated storage budget.

\vspace{-0.5cm}

%\vspace{-0.2cm}
\begin{algorithm}[htbp]
\caption{Over-budget HMN Double-end Pruning}
\label{alg:pruning}
\begin{algorithmic}[1]
\STATE {\bfseries Input:} Over-budget HMN: $H$; Over-budget images per class: $k$; $\beta$ search space $\mathcal{B}$.
% \vspace{0.1cm}
\STATE Condensed dataset $\mathcal{S} \gets H()$; $Acc_{best} = 0$; $\mathcal{S}_{best} = \emptyset$;
\STATE Calculate AUM for all examples in $\mathcal{S}$ based on Equation~\ref{eq:aum};
% \STATE 
\FOR{$\beta$ {\bfseries in} $\mathcal{B}$}
    \STATE $\widetilde{\mathcal{S}} \gets \mathcal{S}.clone()$;
    \STATE Prune $\lfloor \beta k \rfloor$ of the lowest AUM examples for each class from $\widetilde{\mathcal{S}}$;
    \STATE Prune $k - \lfloor \beta k \rfloor$ of the highest AUM examples for each class from $\widetilde{\mathcal{S}}$;
    \STATE Retrain model $f$ on $\widetilde{\mathcal{S}}$;
    \STATE $Acc \gets $ Test accuracy of the model $f$;
    \IF{$Acc > Acc_{best}$}
    \STATE $Acc_{best} = Acc$; $\widetilde{\mathcal{S}}_{best}=\widetilde{\mathcal{S}}$;
    \ENDIF
\ENDFOR
\STATE $\Delta \mathcal{S} = \mathcal{S} - \widetilde{\mathcal{S}}_{best}$;
\STATE $\widetilde{H} \gets$ Prune corresponding instance-level memories in $H$ based on $\Delta \mathcal{S}$;
\STATE {\bfseries Output:} Pruned in-budget network: $\widetilde{H}$.
\end{algorithmic}
\end{algorithm}
%\vspace{-0.4cm}

Inspired by CCS~\cite{zheng2022coverage} showing that pruning both easy and hard data leads to better coreset, we present a double-end pruning algorithm with an adaptive hard pruning rate to prune data adaptively for different storage budgets.
As shown in Algorithm~\ref{alg:pruning}, given an over-budget HMN containing $k$ more generated images per class than allowed by the storage budget, we employ grid search to determine an appropriate hard pruning rate, denoted as $\beta$~(Line 4 to Line 12).  
We then prune $\lfloor \beta k \rfloor$ of the lowest AUM (hardest) examples and $k - \lfloor \beta k \rfloor$ of the highest AUM (easiest) examples by removing the corresponding instance-level memory for each class.
The pruning is always class-balanced: the pruned HMNs generate the same number of examples for each class.

Pruning in Algorithm 1 introduces additional computational costs compared to the standard data condensation pipeline.
However, in practice, the pruning step is relatively cheap (e.g., an additional 2-3\% overhead for data condensation).  %For example, while data condensation with HMNs for CIFAR10 1 IPC needs about 15 hours on a 2080TI GPU, the coreset selection on the condensed dataset only costs an additional 20 minutes.

\vspace{-0.25cm}
\section{Experiments}
\label{sec:experiments}
\vspace{-0.2cm}

In this section, we compare the performance of HMN to SOTA baselines. 
HMN is compared to DC baselines for different values of storage budgets. As with other data parameterization techniques, our condensed data does not store images but rather model parameters. \textbf{All the model parameters of an HMN in Figure~\ref{fig:HMN}, including the three-tier memories and networks, are considered as part of the storage budget.}
For the convenience of comparison, following prior work in the data condensation area~\cite{dengremember, liudataset, kim2022dataset}, the unit used for measuring storage budget is IPC (Images Per Class). 
1 IPC for CIFAR10 is calculated as $32*32*3*1*10 = 30,720$~(assuming that they are stored as 32-bit floating point values). An HMN for CIFAR10 with 1 IPC as the storage budget always has an equal or lower number of parameters than this. 

Due to the page limitation, we include additional evaluation results in Appendix~D. 
Appendix~D.1 examines the relationship between pruning rate and accuracy. 
Appendix~D.2 studies the convergence speed of training for different condensed datasets.
Appendix~D.4 presents results from data profiling to study the data redundancy on the condensed datasets synthesized by different DC methods. 
Finally, Appendix~D.5 presents visualizations of the condensed training data generated by HMNs.

% \ap{I think we can drop the following or rewrite without section numbers.}

%In Section~\ref{ssec:performance-compaison} and discuss the impacts of post-condensation pruning and HMN architecture design in Section~\ref{ssec:ablation}.
%We also compare the transferability of datasets generated by HMNs and other baselines in Section~\ref{ssec:app-transferability}.
%In Section~\ref{ssec:memory-comparision}, we compare the memory usage between HMN and other baselines.
%learning tasks in Section~\ref{ssec:continual-learning}.

% apply HMN to continual learning tasks t

% We also study 
% the cross-architecture transferability of condensed datasets in Section~\ref{ssec:transferability} and their effectiveness in continual learning in 
% \vspace{-0.3cm}
\subsection{Experimental Settings}
\label{ssec:setting}
% \vspace{-0.1cm}

\textbf{Datasets and Training Settings.}
We evaluate our proposed method on four public datasets: CIFAR10, CIFAR100~\cite{krizhevsky2009learning}, SVHN~\cite{svhndataset}, Tiny-ImageNet~\cite{deng2009imagenet}, and ImageNet-10~\cite{deng2009imagenet} under three different storage budgets: 1/10/50 IPC (For Tiny-ImageNet and ImageNet-10, due to the computation limitation, we conduct the evaluation on 1/10 IPC and 1 IPC, respectively).
Following previous works~\cite{zhao2021dataset2, liudataset, dengremember}, we select ConvNet, which contains three convolutional layers followed by a pooling layer, as the network architecture for data condensation and classifier training. 
For the over-budget training and post-condensation, we first conducted a pruning study on HMNs in Appendix~D.1, 
we observed that there is a pronounced decline in accuracy when the pruning rate exceeds 10\%. Consequently, we select 10\% as the over-budget rate for all settings. 
Nevertheless, we believe that this rate choice could be further explored, and other rate values could potentially further enhance the performance of HMNs.
Due to space limits, we include more HMN architecture details, experimental settings, and additional implementation details in the supplementary material.
All data condensation evaluation is repeated $3$ times, and training on each HMN is repeated $10$ times with different random seeds to calculate mean with standard deviation.

\newcommand{\format}[2]{\begin{tabular}{@{}c@{}}$#1$\\[-1.5pt]$\scriptscriptstyle \pm #2$\end{tabular}}
\newcommand{\boformat}[2]{\color{black} \begin{tabular}{@{}c@{}}$\mathbf{#1}$\\[-1.5pt]$\scriptscriptstyle \mathbf{\pm #2}$\end{tabular}}

\begin{table}[!t]
    \centering
    \caption{The performance (test accuracy \%) comparison to state-of-the-art methods. The baseline method's accuracy is obtained from data presented in original papers or author-implemented repos. 
    We label the methods using the trajectory-based training loss with a star~(*). I-10 stands for ImageNet-10.
    We highlight the highest accuracy among all methods and methods with batch-based loss.
    }
    \setlength{\tabcolsep}{2.3pt}
    \scriptsize
    \begin{tabular}{cccccccccccccc}
    \toprule 
    % \multirow{3}{*}{} & Dataset 
    Container &
    \multicolumn{1}{c}{Dataset} & \multicolumn{3}{c}{CIFAR10} & \multicolumn{3}{c}{CIFAR100} & \multicolumn{3}{c}{SVHN} & \multicolumn{2}{c}{Tiny} & I-10\\
    % \midrule \format{}{}
    & \multicolumn{1}{c}{IPC} & 1 & 10 & 50 & 1 & 10 & 50 & 1 & 10 & 50 & 1 & 10 & 1 \\
    \midrule
    \multirow{10}{*}{\begin{tabular}[c]{@{}c@{}}
    Image
    % Loss Function
    \end{tabular}} 
    & DC & \format{28.3}{0.5} &\format{44.9}{0.5} & \format{53.9}{0.5} & \format{12.8}{0.3} & \format{25.2}{0.3} & - & \format{31.2}{1.4} & \format{76.1}{0.6} & \format{82.3}{0.3} & \format{4.6}{0.6} & \format{11.2}{1.6} & - \\
    & DSA & \format{28.8}{0.7} & \format{52.1}{0.5} & \format{60.6}{0.5} & \format{13.9}{0.3} & \format{32.3}{0.3} & \format{42.8}{0.4} & \format{27.5}{1.4} & \format{79.2}{0.5} & \format{84.4}{0.4} & \format{6.6}{0.2} & \format{14.4}{2.0} & -   \\
    & DM & \format{26.0}{0.8} & \format{48.9}{0.6} & \format{63.0}{0.4} & \format{11.4}{0.3} & \format{29.7}{0.3} & \format{43.6}{0.4} & - & - & - & \format{3.9}{0.2} & \format{12.9}{0.4} & - \\
    & CAFE+DSA & \format{31.6}{0.8} & \format{50.9}{0.5} & \format{62.3}{0.4} & \format{14.0}{0.3} & \format{31.5}{0.2} & \format{42.9}{0.2} & \format{42.9}{3.0} & \format{77.9}{0.6} & \format{82.3}{0.4} & - & - & - \\
    
    & MTT* & \format{46.3}{0.8} & \format{65.3}{0.7} & \format{71.6}{0.2} & \format{24.3}{0.3} & \format{40.1}{0.4} & \format{47.7}{0.2} & \format{58.5}{1.4} & \format{70.8}{1.8} & \format{85.7}{0.1} & \format{8.8}{0.3} & \format{23.2}{0.2} & - \\

    & IDM & \format{45.6}{0.7} & \format{58.6}{0.1} & \format{67.5}{0.1} & \format{20.1}{0.3} & \format{45.1}{0.1} & \boformat{50.0}{0.2} & - & - & - & \format{10.1}{0.2} & \format{21.9}{0.2} & - \\

    % & Forgetting & 12.1$\pm$1.7 & 16.8$\pm$1.2 & 27.2$\pm$1.5 & 13.5$\pm$1.2 & 23.3$\pm$1.0 & 23.3$\pm$1.1 & 4.5$\pm$0.3 & 9.8$\pm$0.2 & -  \\
    \midrule
    \multirow{6.5}{*}{\begin{tabular}[c]{@{}c@{}}Data \\ Parame-\\-terization \end{tabular}}
    & IDC & \format{50.0}{0.4} & \format{67.5}{0.5} & \format{74.5}{0.2} & - & $45.1$ & - & $68.5$ & $87.5$ & $90.1$ & - & - & $60.4$ \\
    & HaBa* & \format{48.3}{0.8} & \format{69.9}{0.4} & \format{74.0}{0.2} & \format{33.4}{0.4} & \format{40.2}{0.2} & \format{47.0}{0.2} & \format{69.8}{1.3} & \format{83.2}{0.4} & \format{88.3}{0.1} & - & - & - \\
    & LinBa* &  \boformat{66.4}{0.4} & \format{71.2}{0.4} &  \format{73.6}{0.5} &  \format{34.0}{0.4} &  \format{42.9}{0.7} & - &  \format{87.3}{0.1} &  \format{89.1}{0.2} &  \format{89.5}{0.2} & \format{16.0}{0.7} & - & - \\
    & HMN~(Ours) & \boformat{65.7}{0.3} & \boformat{73.7}{0.2} & \boformat{76.9}{0.2} & \boformat{36.3}{0.2} & \boformat{45.4}{0.2} & \format{48.5}{0.2} & \boformat{87.4}{0.2} & \boformat{90.0}{0.1} & \boformat{91.2}{0.1} & \boformat{19.4}{0.1} & \boformat{24.4}{0.1} & $\mathbf{64.6}$ \\
    \midrule
    % \multicolumn{2}{c}{Random} & $46.3_{\pm0.8}$ & $65.3_{\pm0.7}$ & $71.6_{\pm0.2}$ & $24.3_{\pm0.3}$ & $40.1_{\pm0.4}$ & $47.7_{\pm0.2}$ & $58.5_{\pm1.4}$ & $70.8_{\pm1.8}$ & $85.7_{\pm0.1}$    \\
    \multicolumn{2}{c}{Entire Dataset}  & \multicolumn{3}{c}{$84.8_{\pm0.1}$} & \multicolumn{3}{c}{$56.2_{\pm0.3}$}  & \multicolumn{3}{c}{$95.4_{\pm0.1}$} & \multicolumn{2}{c}{$37.6_{\pm0.4}$} & $90.8$\\
    \bottomrule
    \end{tabular}
\label{tab:compare_sota}
% \end{floatrow}
% \vspace{-0.2cm}
\end{table}

\textbf{Baselines.}
We compare our proposed method with eight baselines, which can be divided into two categories by data containers: 
\textbf{1) Image data container.} 
% Most data condensation methods use image as the container to store condensed information and differs on the training loss to learn condensed images. 
We use five recent works as the baseline: 
MTT~\cite{cazenavette2022dataset} (as mentioned in Section~\ref{sec:related-work}).
DC~\cite{zhao2021dataset} and DSA~\cite{zhao2021dataset2} optimize condensed datasets by minimizing the distance between gradients calculated from a batch of condensed data and a batch of real data. 
DM~\cite{zhao2023dm} aims to encourage condensed data to have a similar distribution to the original dataset in latent space. 
IDM~\cite{zhao2023improved} enhances distribution matching by improving the naive average embedding distribution matching.
Finally, CAFE~\cite{wang2022cafe} improves the distribution matching idea by layer-wise feature alignment.
\textbf{2) Data parameterization.} 
We also compare our method with three SOTA data parameterization baselines.
IDC~\cite{kim2022dataset} enhances gradient matching loss calculation strategy and employs multi-formation functions to parameterize condensed data. HaBa~\cite{liudataset} and 
LinBa~\cite{dengremember} proposes factorization-based data parameterization to achieve information sharing among generated images.
% , which applies  methods to MTT to achieve improved condensation performance; and  LinBa~\cite{dengremember}, which combines a factorization-based data container with backpropagation through time (BPTT).

Besides grouping the methods by data containers, we also categorize those methods by the training losses used.
As discussed in Section~\ref{sec:related-work}, there are two types of training loss: trajectory-based training loss and batch-based training loss. 
% Trajectory-based training losses usually have a better performance than batch-based losses, but need GPU memory to keep track of the long-term training behavior of condensed datasets. 
In Table~\ref{tab:compare_sota}, we highlight the methods using a trajectory-based loss with a star~(*).
In our HMN implementation, we condense HMNs with gradient matching loss used in \cite{kim2022dataset}, which is a low GPU memory consumption batch-based loss.

\vspace{-0.1cm}
\subsection{Data Condensation Performance Comparison}
\vspace{-0.1cm}

\label{ssec:performance-compaison}

We compare HMN with eight baselines on four different datasets (CIFA10, CIFAR100, SVHN, Tiny ImageNet, and ImageNet-10) in Table~\ref{tab:compare_sota}.
We divide all methods into two categories by the type of data container formats: 
Image data container and data parameterization container. 
We also categorize all methods 
by the training loss. 
We use a star~(*) to highlight the methods using a 
trajectory-based loss.
The results presented in Table~\ref{tab:compare_sota} show that HMN achieves comparable or better performance than all baselines.
It is worth noting that HMN is trained with gradient matching, which is a low GPU memory loss. 
Two other well-performed data parameterization methods, HaBa and LinBa, are all trained with trajectory-based losses, consuming much larger GPU memory.
% As far as we know, HMN is the first method to achieve such good performance with a low GPU memory loss.
These results show that batch-based loss can still achieve good performance with an effective data parameterization method and help address the memory issue of data condensation~\cite{cazenavette2022dataset, cuidc, cazenavette2023generalizing}.
We believe that HMN provides a strong baseline for data condensation methods. 
We further study the memory consumed by different methods in Section~\ref{ssec:memory-comparision}.

\begin{wraptable}{r}{0.48\textwidth}
    \centering
    \vspace{-0.3cm}
    \caption{Accuracy~(\%) performance comparison to data containers with the same gradient matching training loss on CIFAR10.
    The evaluation results show that HMN outperforms all other data parameterization methods substantially.}
    \begin{tabular}{c | c c c}
        \hline
        Data Container  & 1 IPC & 10 IPC & 50 IPC\\
        \hline
        Image  & 36.7 & 58.3 & 69.5 \\
        IDC   & 50.0 & 67.5 & 74.5 \\
        HaBa & 48.5 & 61.8 & 72.4 \\
        LinBa  & 62.0 & 67.8 & 70.7 \\
        HMN (Ours) & \textbf{65.7} & \textbf{73.7} & \textbf{76.9} \\
        \hline
    \end{tabular}
    \label{tab:same-loss-comparison}
    \vspace{-0.25cm}
\end{wraptable}
\textbf{Data Parameterization Comparison with the Same Loss.}
In addition to the end-to-end method comparison presented in Table~\ref{tab:compare_sota}, 
we also compare HMN with other data parameterization methods with the same training loss (gradient matching loss used by IDC) for a fairer comparison.
The results are presented in Table~\ref{tab:same-loss-comparison}.
After replacing the trajectory-based loss used by HaBa and LinBa with a batch-based loss, there is a noticeable decline in accuracy (but HaBa and LinBa still outperform the image data container).\footnote{We do hyperparameter search for all data containers to choose the optimal setting.}
HMN outperforms other data parameterization by a larger margin when training with the same training loss, which indicates that HMN is a more effective data parameterization method and can condense more information within the same storage budget.
We also discussed the memory consumption of trajectory-based methods in Section~\ref{ssec:memory-comparision}.

\vspace{-0.3cm}
\subsection{Cross-architecture Transferability} 
\label{ssec:app-transferability}
% \vspace{-0.6cm}

\begin{table}[htbp]
    \centering
    \caption{Transferability (accuracy \%) comparison to different model architectures. Due to the extremely long training time, we cannot reproduce the results on LinBa. Compared with IDC and HaBa, we find that HMN achieves better performance for all model architectures. }
\begin{tabular}{c|ccc|ccc|ccc}
        \hline
        IPC &  \multicolumn{3}{c|}{1} &\multicolumn{3}{c|}{10} & \multicolumn{3}{c}{50}\\
        \hline
        Method &  HMN & IDC & HaBa & HMN & IDC & HaBa & HMN & IDC & HaBa\\
        \hline
        ConvNet &  \textbf{65.7} & 50.0 & 48.3 & \textbf{73.7} & 67.5 & 69.9 & \textbf{76.9} & 74.5 & 74\\
        VGG16 &  \textbf{58.5} & 28.7 & 34.1 & \textbf{64.3} & 43.1 & 53.8 & \textbf{70.2} & 57.9 & 61.1 \\
        ResNet18 &  \textbf{56.8} & 32.3 & 36.0 & \textbf{62.9} & 45.1 & 49.0 & \textbf{69.1} & 58.4 & 60.4\\
        DenseNet121 &  \textbf{50.7} & 24.3 & 34.6 & \textbf{56.9} & 38.5 & 49.3 & \textbf{65.1} & 50.5 & 57.8\\
        \hline
\end{tabular}
\vspace{-0.4cm}
\label{tab:transferability}
\end{table}

To investigate the generalizability of HMNs across different architectures, we utilized condensed HMNs to train other network architectures. 
Specifically, we condense HMNs with ConvNet, and the condensed HMNs are tested on VGG16, ResNet18, and DenseNet121.
We compare our methods with two other data parameterization methods: IDC and HaBa. (Due to the extremely long training time, we are unable to reproduce the results in LinBa).
The evaluation results on CIFAR10 are presented in Table~\ref{tab:transferability}.
We find that HMNs consistently outperform other baselines.
Of particular interest, we observe that VGG16 has a better performance than ResNet18 and DenseNet121.
A potential explanation may lie in the architectural similarities between ConvNet and VGG16. Both architectures are primarily comprised of convolutional layers and lack skip connections.

\vspace{-0.3cm}
\subsection{GPU Memory Comparison}
\label{ssec:memory-comparision}
\vspace{-0.15cm}

As discussed in Section~\ref{ssec:setting}, GPU memory consumption can be very different depending on the training losses used.
We compare the GPU memory used by HMN with two other well-performed data parameterization methods, Ha- Ba and LinBa.
As depicted by Table~\ref{tab:memory-comparison}, HMN achieves better or comparable 
\begin{wraptable}{r}{0.6\textwidth}
    \centering
    \vspace{-0.7cm}
    \setlength{\tabcolsep}{3.2pt}
    % \scriptsize
    \caption{Performance and memory comparison between LinBa, HaBa, and HMN trained with suggested loss in corresponding papers on CIFAR10. OOM means  "Out of GPU Memory." Reported accuracy numbers use CPU offloading in OOM case.}
    \begin{tabular}{c | c | c c c}
        \hline
        IPC &  Method & Loss & Acc. & Memory\\
        \hline
            \multirow{3}{*}{\begin{tabular}[c]{@{}c@{}}
            1
            % Loss Function
            \end{tabular}} 
        & HaBa & MTT & 48.3 & 3368M \\
        & LinBa & BPTT & \textbf{66.4} & OOM \\
         & HMN (Ours) & GM-IDC & 65.7 & \textbf{2680M} \\
        \hline
            \multirow{3}{*}{\begin{tabular}[c]{@{}c@{}}
            10
            % Loss Function
            \end{tabular}} 
          & HaBa & MTT & 69.9 & 11148M \\
        & LinBa & BPTT & 71.2 & OOM \\
         & HMN (Ours) & GM-IDC & \textbf{73.7} & \textbf{4540M} \\
        \hline
            \multirow{3}{*}{\begin{tabular}[c]{@{}c@{}}
            50
            % Loss Function
            \end{tabular}} 
          & HaBa & MTT & 74 & 48276M \\
          & LinBa & BPTT & 73.6 & OOM \\
          & HMN (Ours) & GM-IDC & \textbf{76.9} & \textbf{10426M} \\
        \hline
    \end{tabular}
    % \vspace{-0.3cm}
    \label{tab:memory-comparison}
\end{wraptable}
performance compared to HaBa and LinBa with much less GPU memory consumption. 
Specifically, LinBa is trained with BPTT with a very long trajectory, which leads to extremely large GPU memory consumption. 
LinBa official implementation offloads the GPU memory to CPU memory to address this issue. 
However, the context switch in memory offloading causes the training time to be intolerable. For example, LinBa needs about 14 days to condense a CIFAR10 1IPC dataset with a 2080TI, but using HMN with gradient matching only needs 15 hours to complete training on a 2080TI GPU.
Although this memory saving does not come from the design of HMN, our paper shows that batch-based loss can still achieve very good performance with a proper data parameterization method, which helps address the memory issue of data condensation~\cite{cazenavette2022dataset, cuidc, cazenavette2023generalizing}.

% \ap{Acknowledge that combining HMN with trajectory-based loss is too expensive on the GPUs we have.}

Combining an HMN with the trajectory-based loss may further improve the performance of an HMN-based approach, but the cost is too high for the GPUs we have.
We leave that investigation to future work since using trajectory-based loss in practice remains challenging from a scalability perspective due to high memory needs. 
For a fair comparison, we compare the different data containers with the same batch-based loss in Table~\ref{tab:same-loss-comparison}.

% \vspace{-0.5cm}
\subsection{Ablation Studies} 
\label{ssec:ablation}
% \vspace{-0.05cm}

\textbf{Instance-Memory Size v.s. Retrained Model Accuracy.} 
In an HMN, every generated image is associated with an independent instance-level memory, which constitutes the majority of the storage budget. Consequently, given a fixed storage budget, an increase in the instance-level memory results in a decrease in the number of generated images per class (GIPC).
In Figure~\ref{fig:instance-memory}, we explore the interplay between the instance-memory size, the accuracy of the retrained model, and GIPC. Specifically, we modify the instance-level memory size of CIFAR10 HMNs for given storage budgets of 1 IPC and 10 IPC. (It should be noted that for this ablation study, we are condensing in-budget HMNs directly without employing any coreset selection on the condensed HMNs.)

From Figure~\ref{fig:instance-memory}, we observe that an increase in the instance-level memory size leads to a swift drop in GIPC, as each generated image consumes a larger portion of the storage budget. 
Moreover, we notice that both excessively small and large instance-level memory sizes negatively affect the accuracy of retrained models.
Reduced instance-level memory size can result in each generated image encoding only a limited amount of information. This constraint can potentially deteriorate the quality of the generated images and negatively impact training performance. Conversely, while an enlarged instance-level memory size enhances the volume of information encoded in each image, it precipitously reduces GIPC. This reduction can compromise the diversity of generated images for training.
For instance, with a 1IPC storage budget, an increase in the instance-level memory size, leading to a decrease in GIPC from $85$ to $13$, results in an accuracy drop from $65.1\%$ to $48.2\%$.

\begin{figure}[t]
  \centering
  \begin{minipage}{0.48\textwidth}
    \centering
    \includegraphics[width=1\textwidth]{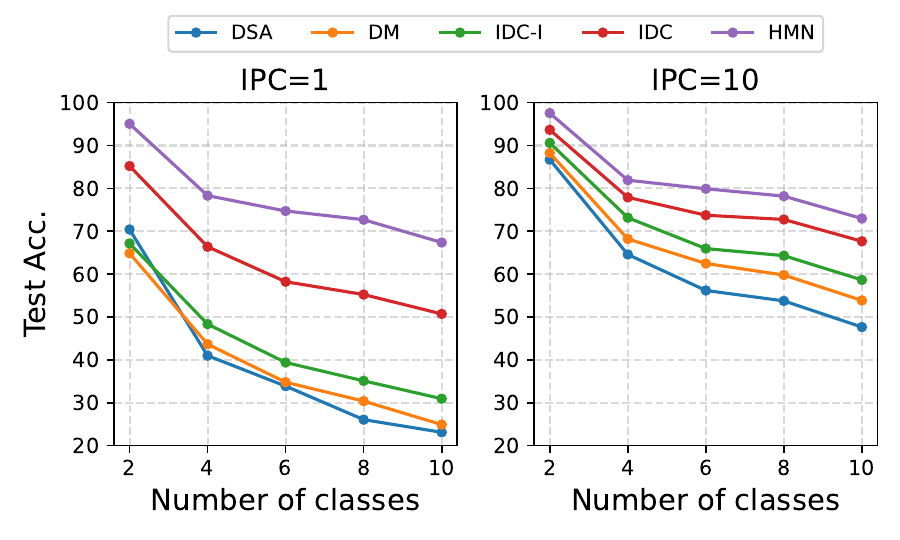}
    \caption{Continual learning evaluation on CIFAR10. In the class incremental setting with 2 incoming classes per stage, HMN outperforms existing methods (including DSA, DM and IDC) under different storage budgets. }
    % \caption{Instance-memory length vs. Retrain Accuracy for CIFAR-10 HMNs with a size budget of 1IPC and 10IPC. GIPC refers to the number of generated images per class. The solid curve represents accuracy, while the dashed curve represents GIPC.}
    \label{fig:continual}
  \end{minipage}\hfill
  \begin{minipage}{0.48\textwidth}
    \centering
    \includegraphics[width=1\textwidth]{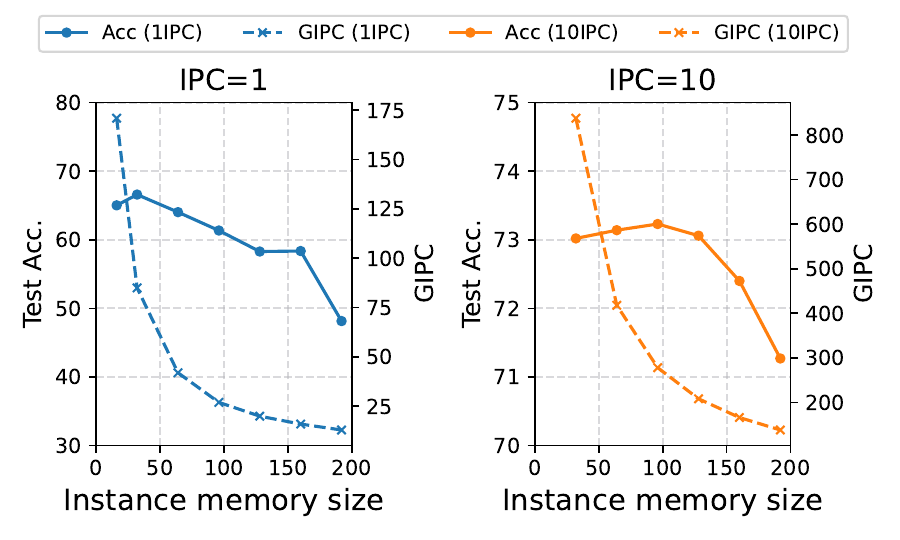}
    \caption{Instance-memory length vs. Accuracy for CIFAR10 HMNs with 1 IPC/10 IPC storage budgets. GIPC refers to the number of generated images per class. The solid and dashed curves represent the accuracy and GIPC, respectively.}
    \label{fig:instance-memory}
  \end{minipage}
  % \vspace{-0.15cm}
\end{figure}
% A small instance-level memory size can lead to each generated image only encoding limited information, which potentially hurt the generated image quality and hurt the training performance. 
% On the other side, although a large instance-level memory size can improve the amount of information encoded in each image, it drastically decreases GIPC, which can hurt the diversity of generated images for training.
% For example, given an 1IPC size budget, when we decrease GPIC from $85$ to $13$ by increasing the instance-level memory size, the accuracy drops from $65.1\%$ to $48.2\%$.

\textbf{Ablation Study on Pruning.} In Table~\ref{tab:pruning-ablation}, we explore the performance of different pruning strategies applied to over-budget HMNs on the CIFAR10 dataset. 
The strategy termed "Prune easy" is widely employed in conventional coreset selection methods~\cite{coleman2019selection, toneva2018empirical, paul2021deep, xiamoderate}, which typically prioritize pruning of easy examples containing more redundant information.
``In-budget" refers to the process of directly condensing HMNs to fit the storage budgets, which does not need any further pruning.
As shown in Table~\ref{tab:pruning-ablation}, our proposed pruning strategy (double-end) 
outperforms all other pruning strategies. 
We also observe that, as the storage 
\begin{wraptable}{r}{0.45\textwidth}
    \vspace{-0.7cm}
    \centering
    \caption{Performance comparison on different pruning strategies on HMN. Double-end is the pruning strategy introduced in Algorithm~\ref{alg:pruning}. In-budget stands for HMNs are condensed within the allocated storage budget.}
    \begin{tabular}{cccc}
    \toprule
    \multicolumn{1}{c}{IPC} & 1 & 10 & 50\\
    \midrule
    Double-end & $\mathbf{65.7}$ & $\mathbf{73.7}$ & $\mathbf{76.9}$\\
    Prune easy & $65.3$ & $73.1$ & $76.6$ \\
    Random & $65.2$ & $72.9$ & $75.3$ \\
    In-budget & $65.1$ & $73.2$ & $75.4$ \\
    \bottomrule
    \end{tabular}
    \label{tab:pruning-ablation}
    \vspace{-0.3cm}
\end{wraptable}
budget increases, the accuracy improvement becomes larger compared to ``in-budget'' HMNs.
We think this improvement is because a larger storage budget causes more redundancy in the condensed data~\cite{cuidc}, 
which makes pruning reduce more redundancy in condensed datasets.
Also, the performance gap between the ``Prune easy" strategy and our pruning method is observed to narrow as the storage budget increases. This may be attributed to larger storage budgets for HMNs leading to more redundant easy examples. 
The ``Prune easy" is a good alternative for pruning for large budgets.

\vspace{-0.2cm}
\subsection{Continual Learning Performance Comparison} 
\label{ssec:continual-learning}

% [placeholder] Continual learning 
% \hz{leave for Shutong}

% Continual learning 
Following the same setting in DM~\cite{zhao2023dataset} and IDC~\cite{kim2022dataset},
we evaluate the effectiveness of HMN in an application scenario of continual learning~\cite{bang2021rainbow, rebuffi2017icarl, chaudhry2019tiny}.
Specifically, we split the whole training phase into 5 stages, \emph{i.e.} 2 classes per stage. 
At each stage, we condense the data currently available at this stage with ConvNet.
As illustrated in Figure \ref{fig:continual}, evaluated on ConvNet models under the storage budget of both 1 IPC and 10 IPC, HMN obtains better performance compared with DSA \cite{zhao2021dataset2}, DM \cite{zhao2023dm}, and IDC \cite{kim2022dataset}. 
Particularly, in the low storage budget scenario, \emph{i.e.} 1 IPC, the performance improvement brought by HMN is more significant, up to 16\%. 
The results indicate that HMNs provide higher-quality condensed data and boost continual learning performance.

% \subsection{Additional Results}

% One significant resource advantage of HMNs is that even when using a low-memory, batch-based loss for optimization, HMNs can outperform current trajectory-loss methods that require high GPU memory and thus are prone to potential scalability issues~\cite{cazenavette2022dataset,cuidc}.

%While data parameterization methods demonstrate effective performance in data condensation, we show that generated images per class (GIPC) plays an important role in data parameterization,  Given the same training loss, the computational cost is proportional to the number of generated images.  This characteristic can result in data parameterization approaches being more computationally demanding, a limitation prevalent across all such methods.

\vspace{-0.3cm}
\section{Conclusion}
\vspace{-0.1cm}

This paper introduces a novel data parameterization architecture, Hierarchical Memory Network (HMN), which is inspired by the hierarchical nature of common feature sharing in datasets.
% Inspired by the hierarchical nature of common feature sharing in datasets, we introduce a novel data parameterization architecture, Hierarchical Memory Network (HMN). 
In contrast to previous data parameterization methods, HMN aligns more closely with this hierarchical nature of datasets.
% \ap{Remove redundancy. Add some punchlines on key contributions and perhaps more specifics on memory needs and performance results.}
% Additionally, we also show that redundant data exists in condensed datasets. 
% Unlike previous data parameterization methods, 
% although HMNs achieve information sharing among generated images, 
% HMNs also naturally ensure good independence between generated images, which facilitates the pruning of data containers.
% The evaluation results on five public datasets show that HMN outperforms DC baselines, indicating that HMN is a more efficient architecture for DC.
Our evaluation results show that the proposed HMN data container architecture, even when employing a batch-based loss for its optimization,  achieves better or comparable performance than SOTA DC baselines, including those that utilize high-memory trajectory-based loss functions.

% A contribution of our work is a finding that, by using HMN architecture as a data container, a batch-based loss can achieve comparable and even better performance than current data container methods that are based on memory-intensive trajectory-based loss.

%We propose a novel prunable data container, Hierarchical Memory Network (HMN), along with a unique algorithm specifically designed for coreset selection within HMNs. Contrary to SOTA factorization-based data parameterization approaches, the design of HMNs is closer to classic neural networks, offering better flexibility to architect more adaptive structures tailored to different data condensation scenarios. HMNs also address the interdependence issue of SOTA factorization-based methods, which permits the pruning of individual examples in an independent manner, a notable improvement over the current methods. 

\section*{Acknowledgement}
This work was partially supported by DARPA under agreement number 885000 and National Science Foundation Grant No. 2039445. This work was performed under the auspices of the U.S. Department of Energy by the Lawrence Livermore National Laboratory under Contract No. DE- AC52-07NA27344 and was supported by the LLNL-LDRD Program under Project No. 22-SI-004 and 24-ERD-010. Any opinion, findings, and conclusions or recommendations expressed in this material are those of the authors(s) and do not necessarily reflect the views of our research sponsors.

% \clearpage

% \newpage
% \bibliography{iclr2024_conference}
% \bibliographystyle{iclr2024_conference}

%
\bibliographystyle{splncs04}
\bibliography{main}

\appendix

\section{Appendix Overview}
In this appendix, we provide more details on our experimental setting and additional evaluation results. 
In Section~\ref{sec:app-discussion}, we discuss the computation cost caused by data parameterization.
In Section~\ref{sec:app-setting}, we introduce the detailed setup of our experiment to improve the reproducibility of our paper.
In Section~\ref{sec:app-eval}, we conduct additional studies on how the pruning rate influences the model performance.
In addition, we visualize images generated by HMNs in Section~\ref{ssec:app-visualization}.
% We also include our code in the supplementary material.

\section{Discussion}
\label{sec:app-discussion}
% \section{Limitations}

While data parameterization methods demonstrate effective performance in data condensation, we show that generated images per class (GIPC) play an important role in data parameterization.
The payoff is that HMNs, along with other SOTA data parameterization methods~\cite{kim2022dataset, dengremember,liudataset} invariably generate a higher quantity of images than those condensed and stored in pixel space with a specific storage budget, which may potentially escalate the cost of data condensation.
A limitation of HMNs and other data parameterization methods is that determining the parameters of the data container to achieve high-quality data condensation can be computationally demanding.
Besides, more generated images can lead to longer training time with condensed datasets.
In Section~\ref{ssec:training-time-comparison}, we show that, even though HMNs generate more training images, training on condensed datasets generated by HMNs achieves better test accuracy within the same training time.
% A key challenge and promising future direction is to explore approaches to reduce GIPC without affecting DC performance.

Another difference between data parameterization and conventional DC methods using images as data containers is that data parameterization methods need to generate images before training with condensed datasets. 
It is important to note that this additional step incurs only a minimal overhead, as it merely requires a single forward pass of HMNs. 
For example, on a 2080TI, the generation time for a 1 IPC, 10 IPC, and 50 IPC CIFAR10 HMN is 0.036s, 0.11s, and 0.52s, respectively (average number through 100 repeats).
\section{Experiment Setting and Implementation Details}
\label{sec:app-setting}
% In this section, we introduced more experimental setups and implementation details on the evaluation that we conducted in the main paper.

\subsection{HMN architecture design.}

In this section, we introduce more details on the designs of the Hierarchical Memory Network (HMN) architecture, specifically tailored for various datasets and storage budgets. We first introduce the three-tier hierarchical memories incorporated within the network. Subsequently, we present the neural network designed to convert memory and decode memories into images utilized for training.

% Here we introduce the detailed Hierarchical
% Memory Network~(HMN) architecture design for different datasets and storage budgets. 
% We first introduce the tree-tier hierarchical memories in the network and then introduce the neural network to covert memory and decode memories to images used for training.

\begin{table}[H]
    \centering
    \caption{The detailed three-tier memory settings. We use the same setting for CIFAR10 and SVHN. \#Instance-level memory is the number of memory fitting the storage budget. \#Instance-level memory (Over-budget) indicates the actual number of instance-level memory that we use for condensation, and we prune this number to \#Instance-level memory after condensation. I-10 stands for ImageNet-10.}
    \setlength{\tabcolsep}{2.8pt}
    \scriptsize
    \begin{tabular}{ccccccccccc}
    \toprule
    \multicolumn{1}{c}{Dataset} & \multicolumn{3}{c}{SVHN \& CIFAR10} & \multicolumn{3}{c}{CIFAR100} & \multicolumn{2}{c}{Tiny } & I-10\\
    \multicolumn{1}{c}{IPC} & 1 & 10 & 50 & 1 & 10 & 50 & 1 & 10 & 1\\
    \midrule
    Dataset-level memory channels & 5 & 50 & 50 & 5 & 50 & 50 & 30 & 50  & 30\\
    Class-level memory channels & 3 & 30 & 30 & 3 & 30 & 30 & 20 & 30 & 25\\
    Instance-level memory channels & 2 & 6 & 8 & 2 & 8 & 14 & 4 & 10 & 8\\
    \#Instance-level memory & 85 & 278 & 1168 & 93 & 219 & 673 & 42 & 185 & 125 \\
    \#Instance-level memory (Over-budget) & 93 & 306 & 1284 & 102 & 243 & 740 & 46 & 203 & 138\\
    \bottomrule
    \end{tabular}
\label{tab:setting-memory}
\end{table}

\textbf{Hierarchical memories.}
HMNs consist of three-tier memories: dataset-level memory $m^{(D)}$, class-level memory $m^{(C)}_c$, and instance-level memory $m^{(I)}_{c, i}$, which are supposed to store different levels of features of datasets.
Memories of HMNs for SVHN, CIFAR10, and CIFAR100 have a shape of (4, 4, Channels), and memories for Tiny ImageNet have a shape of (8, 8, Channels).
Memories for ImageNet-10 have a shape of (12, 12, Channels).
The number of channels is a hyper-parameter for different settings.

We present the detailed setting for the number of channels and the number of memories under different data condensation scenarios in Table~\ref{tab:setting-memory}.
Besides the channels of memories, we also present the number of instance-level memories. 
Since each instance-level memory corresponds to a generated image, 
the number of instance-level of memories is the GPIC for an HMN.
Every HMN has only one dataset-level memory, and the number of class-level memory is equal to the number of classes in the dataset.
The number of instance-level memory for the over-budget class leads to an extra 10\% storage budget cost, which will be pruned by post-condensation pruning.
% \hz{We may wanna have some discussion on this part.}

\textbf{Decoders.}
In addition to three-tier memories, each HMN has two types of networks: 1) A dataset-level memory feature extractor for each class; 2) A uniform decoder to convert memories to images for model training.
\emph{Dataset-level memory feature extractors} $f_c$ are used to extract features from the dataset-level memory for each class. 
For 1 IPC storage budget setting, we use the identity function as the feature extractor to save the storage budget. For 10 IPC and 50 IPC storage budget settings, the feature extractors consist of a single deconvolutional layer with the kernel with 1 kernel size and $40$ output channels.
\emph{The uniform decoder} $D$ is used to generate images for training.
For ImageNet-10, the size of the generated image is (3, 96, 96). We use the bilinear interpolation to resize the generated images to (3, 224, 224).
In this paper, we adopt a classic design of decoder for image generation, which consist of a series of deconvolutional layers and batch normalization layers:
ConvTranspose(Channels of memory, 10, 4, 1, 2) $\to$ Batch Normalization $\to$ ConvTranspose(10, 6, 4, 1, 2) $\to$ Batch Normalization $\to$ ConvTranspose(6, 3, 4, 1, 2). The arguments for ConvTranspose is input-channels, output-channels, kernel size, padding, and stride, respectively. The ``Channels of memory'' is equal to the addition of the channels of the output of $f_c$, the class-level memory channels, and the instance-level memory channels. 
When we design the HMN architecture, we also tried the design with different decoders for different classes. However, we find that it experiences an overfitting issue and leads to worse empirical performance.

\subsection{Training settings}
\label{app:exp-setting}

\textbf{Baseline Settings}
In this paper, we evaluate HMN on the same model and architecture and with the same IPC setting as the baselines for a fair comparison. For various baselines, we directly report the numbers represented in their papers.
% use the hyperparameter settings that were recommended in the corresponding papers or used in available code.
In general, as far as we can tell, the authors of various baselines chose reasonable hyperparameter settings, such as learning rate, learning rate schedule, batch size, etc. for their scheme. Sometimes the chosen settings differ. For instance,  LinBa~\cite{dengremember} uses 0.1 as the learning rate, but HaBa~\cite{liudataset} uses 0.01 as the learning rate. In keeping with past work in this area, we accept such differences, since the goal of each scheme is to achieve the best accuracy for a given IPC setting. The settings that we found to be reasonable choices for HMN are  described below. The metrics on which all schemes are being evaluated are the same: accuracy that the scheme is able to achieve for a given IPC setting.

\textbf{Data condensation.}
We generally follow the guidance and settings from past work for the data condensation component of HMN. Following previous works~\cite{zhao2021dataset2, liudataset, dengremember}, we select ConvNet, which contains three convolutional layers followed by a pooling layer, as the network architecture for data condensation and classifier training for all three datasets.
For ImageNet-10, following previous work, we choose ResNet-AP (a four-layer ResNet) to condense HMNs.
We employ gradient matching~\cite{kim2022dataset, liudataset}, a batch-based loss with low GPU memory consumption, to condense information into HMNs.
More specifically, our code is implemented based on IDC~\cite{liudataset}.
For all datasets, we set the number of inner iterations to 200 for gradient matching loss.
The total number of training epochs for data condensation is 1000.
We use the Adam optimizer ($\beta_1= 0.9$ and $\beta_2= 0.99$) with a 0.01 initial learning rate (0.02 initial learning rate for CIFAR100) for data condensation.
The learning rate scheduler is the step learning rate scheduler, and the learning rate will time a factor of $0.1$ at $600$ and $800$ epochs.
We use the mean squared error loss for calculating the distance of gradients for CIFAR10 and SVHN, and use L1 loss for the CIFAR100 and Tiny ImageNet.
To find the best hard pruning rate $\beta$ in Algorithm~1, we perform a grid search from 0 to 0.9 with a 0.1 step.
All experiments are run on a combination of RTX2080TI, RTX3090, A40, and A100, depending on memory usage and availability.

\textbf{Model training with HMNs.} 
For CIFAR10, we train the model with datasets generated by HMNs for 2000, 2000, and 1000 epochs for 1 IPC, 10 IPC, and 50 IPC, respectively.
We use the SGD optimizer (0.9 momentum and 0.0002 weight decay) with a 0.01 initial learning rate.
% The learning rate scheduler is the cosine annealing learning rate scheduler~\cite{loshchilov2016sgdr} with a $0.0001$ minimum learning rate.

For CIFAR100, we train the model with datasets generated by HMNs for 500 epochs.
We use the SGD optimizer ($0.9$ momentum and $0.0002$ weight decay) with a $0.01$ initial learning rate. 
% The learning rate scheduler is the cosine annealing learning rate scheduler~\cite{loshchilov2016sgdr} with a $0.0001$ minimum learning rate.

For SVHN, we train the model with datasets generated by HMNs for 1500, 1500, 700 epochs for 1 IPC, 10 IPC, and 50 IPC, respectively.
We use the SGD optimizer (0.9 momentum and 0.0002 weight decay) with a 0.01 initial learning rate.

For both Tiny-ImageNet and ImageNet-10, we train the model with datasets generated by HMNs for 300 epochs for both 1 IPC and 10 IPC settings.
We use the SGD optimizer (0.9 momentum and 0.0002 weight decay) with a 0.02 initial learning rate.

Similar to \cite{liudataset}, we use the DSA augmentation~\cite{zhao2021dataset2} and CutMix as data augmentation for data condensation and model training on HMNs.
For HMN, for the learning rate scheduler, we use  the cosine annealing learning rate scheduler~\cite{loshchilov2016sgdr} with a $0.0001$ minimum learning rate.
We preferred it over the multi-step learning rate scheduler primarily because the cosine annealing learning rate scheduler has fewer hyperparameters to choose. We also did an ablation study on the learning rate scheduler choice (see Appendix~\ref{ssec:ablation-lr-scheduler}) and did not find the choice of the learning rate scheduler to have a significant impact on the performance results.

\textbf{Continual learning.}
Following the class incremental setting of \cite{kim2022dataset}, we adopt distillation loss \cite{8107520} and train the model constantly by loading weights of the
previous stage and expanding the output dimension of the last fully-connected layer \cite{Rebuffi_2017_CVPR}. Specifically, we use a ConvNet-3 model trained for 1000 epochs at each stage, using SGD with a momentum of 0.9 and a weight decay of $5e-4$. The learning rate is set to 0.01, and decays at epoch 600 and 800, with a decaying factor of 0.2.

% \newpage

\section{Additional Evaluation Results}
\label{sec:app-eval}

In this section, we present additional evaluation results to further demonstrate the efficacy of HMNs. 
% We compare the transferability of datasets generated by HMNs and other baselines in Section~\ref{ssec:app-transferability}.
We study the relationship between pruning rate and accuracy in Section~\ref{ssec:app-pruning-rate-acc}.
We then compare the training time with the condensed datasets in Section~\ref{ssec:training-time-comparison}.
Subsequently, we conduct an ablation study on how different learning rate scheduler influences the training on condensed datasets in Section~\ref{ssec:ablation-lr-scheduler}.
Additionally, we do data profiling and study the data redundancy on the condensed datasets synthesized by different DC methods in Section~\ref{ssec:app-profiling}.
Lastly, we visualize the condensed training data generated by HMNs for different datasets in Section~\ref{ssec:app-visualization}.

% \subsection{Evaluation on ImageNet-10} 
% \label{ssec:app-imageNet-10}
% To investigate 

% \begin{wraptable}{r}{0.45\textwidth}
%     % \vspace{-0.2cm}
%     \centering
%     \caption{Performance comparison on ImagNet-10.}
%     \begin{tabular}{cccc}
%     \toprule
%     \multicolumn{1}{c}{IPC} & 1 & 10 & 50\\
%     \midrule
%     Double-end & $\mathbf{65.7}$ & $\mathbf{73.7}$ & $\mathbf{76.9}$\\
%     Prune easy & $65.3$ & $73.1$ & $76.6$ \\
%     Random & $65.2$ & $72.9$ & $75.3$ \\
%     In-budget & $65.1$ & $73.2$ & $75.4$ \\
%     \bottomrule
%     \end{tabular}
%     \label{tab:pruning-ablation}
% \end{wraptable}

\subsection{Pruning Rate v.s. Accuracy}
\label{ssec:app-pruning-rate-acc}

In this section, we examine the correlation between accuracy and pruning rates on HMNs.
The evaluation results are presented in Figure~\ref{fig:pruning-rate}. 
We observe that the accuracy drops more as the pruning rates increase, and our double-end pruning algorithm consistently outperforms random pruning. 
Furthermore, we observe that an increasing pruning rate results in a greater reduction in accuracy for HMNs with smaller storage budgets.
For instance, when the pruning rate increases from 0 to 30\%, models trained on the 1 IPC HMN experience a significant drop in accuracy, plunging from 66.2\% to 62.2\%. 
Conversely, models trained on the 50 IPC HMN exhibit a mere marginal decrease in accuracy, descending from 76.7\% to 76.5\% with the same increase in pruning rate. 
This discrepancy may be attributed to the fact that HMNs with larger storage budgets generate considerably more redundant data. Consequently, pruning such data does not significantly impair the training performance.

\subsection{Training Time Comparison with Condensed Datasets}
\label{ssec:training-time-comparison}

\begin{figure}[h]
    \centering
    \vspace{0pt}
    \includegraphics[width=\linewidth]{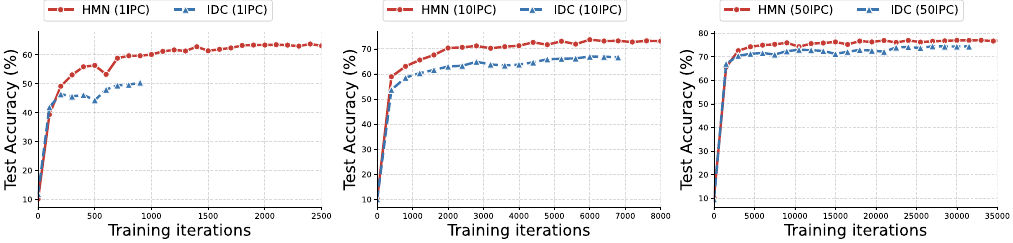}
    \caption{Test accuracy over training iterations of training with condensed datasets generated by HMN and IDC. Given the same storage budget, although training with HMN needs more training iterations to converge, we find that HMNs achieve better accuracy within the same training time compared to IDC.}
    \label{fig:training-time}
    % \vspace{-0.3cm}
\end{figure}

One potential limitation of HMNs is that, given the same storage budget, HMNs generate more images than other DC methods, which can potentially increase the cost of training with condensed datasets generated by HMNs.
In this section, we conduct a study to study how test accuracy changes with respect to training iterations. We use the same batch size for both methods and follow the training setting suggested in the IDC paper.
The comparison results are illustrated in Figure~\ref{fig:training-time}.
Although condensed datasets generated by HMNs contain more training images, training with HMNs achieves better accuracy within the same training time across different training budgets.
For instance, for 1 IPC at the 900th iteration, HMN achieves an accuracy of 60.1\% while IDC only achieves 50.4\% (at this point, IDC has converged, while HMN's accuracy can still be boosted further with more training iterations). 

\subsection{Ablation Study on Learning Rate Scheduler}
\label{ssec:ablation-lr-scheduler}

\begin{wraptable}{r}{0.5\textwidth}
\vspace{-0.3cm}
    \centering
    \caption{
    Accuracy~(\%) performance comparison on different LR scheduler on CIFAR10.
    The evaluation results show that the difference due to the LR scheduler choice is overall marginal.
    }
    \begin{tabular}{c | c c c}
        \hline
        Data Container  & 1 IPC & 10 IPC & 50 IPC\\
        \hline
        Multi-step  & 65.7 & 73.4 & 76.8 \\
        Cosine Annealing   & 65.7 & 73.7 & 76.9 \\
        \hline
    \end{tabular}
    \label{tab:ablation-lr}
    \vspace{-0.15cm}
\end{wraptable}

% {\color{blue}
We also train the model with a multi-step learning rate scheduler on CIFAR10 datasets generated by HMNs and found the following hyperparameter settings for a multi-step learning rate scheduler to work well: (a) an initial learning rate of 0.1; (b) The learning rate is multiplied with a 0.1 learning rate decay at 0.3 * total epochs / 0.6 * total epochs /  0.9 * total epochs.   As shown in Table~\ref{tab:ablation-lr}, we find the difference due to the LR scheduler choice to be overall marginal, and the results with the multistep LR scheduler do not change the findings of our evaluation. Our primary reason for choosing the cosine annealing LR scheduler in our evaluation is that it has fewer hyperparameters to choose from compared to the multistep LR scheduler. The cosine annealing LR scheduler only requires selection of an initial learning rate and a minimum learning rate. Those settings are described in Appendix~\ref{app:exp-setting}.

% }

% \subsection{More transferability evaluation}

\subsection{Data Profiling on SOTA Methods}
\label{ssec:app-profiling}

\begin{figure}[htbp]
  \centering
  \begin{minipage}{0.48\textwidth}
    \centering
    \includegraphics[width=1\textwidth]{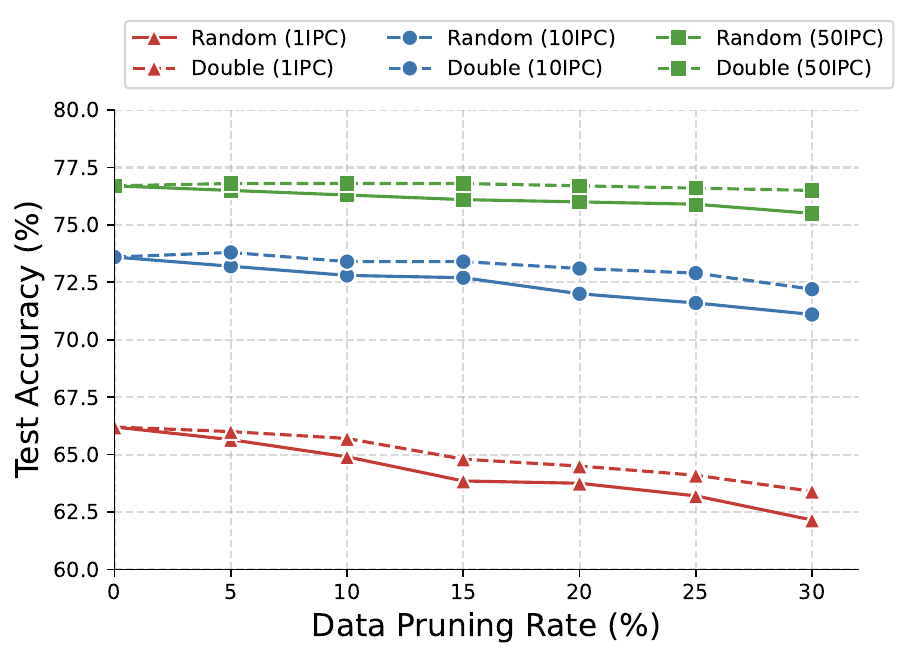}
    \caption{Relationship between pruning rates and accuracy on HMNs for CIFAR10. All HMNs are over-budget HMNs (10\% extra). 
    Different colors stand for different storage budgets.
    Solid lines stand for random pruning and dashed lines stand for double-end pruning.
    }
    \label{fig:pruning-rate}
  \end{minipage}\hfill
  \begin{minipage}{0.48\textwidth}
    \centering
    \includegraphics[width=1\textwidth]{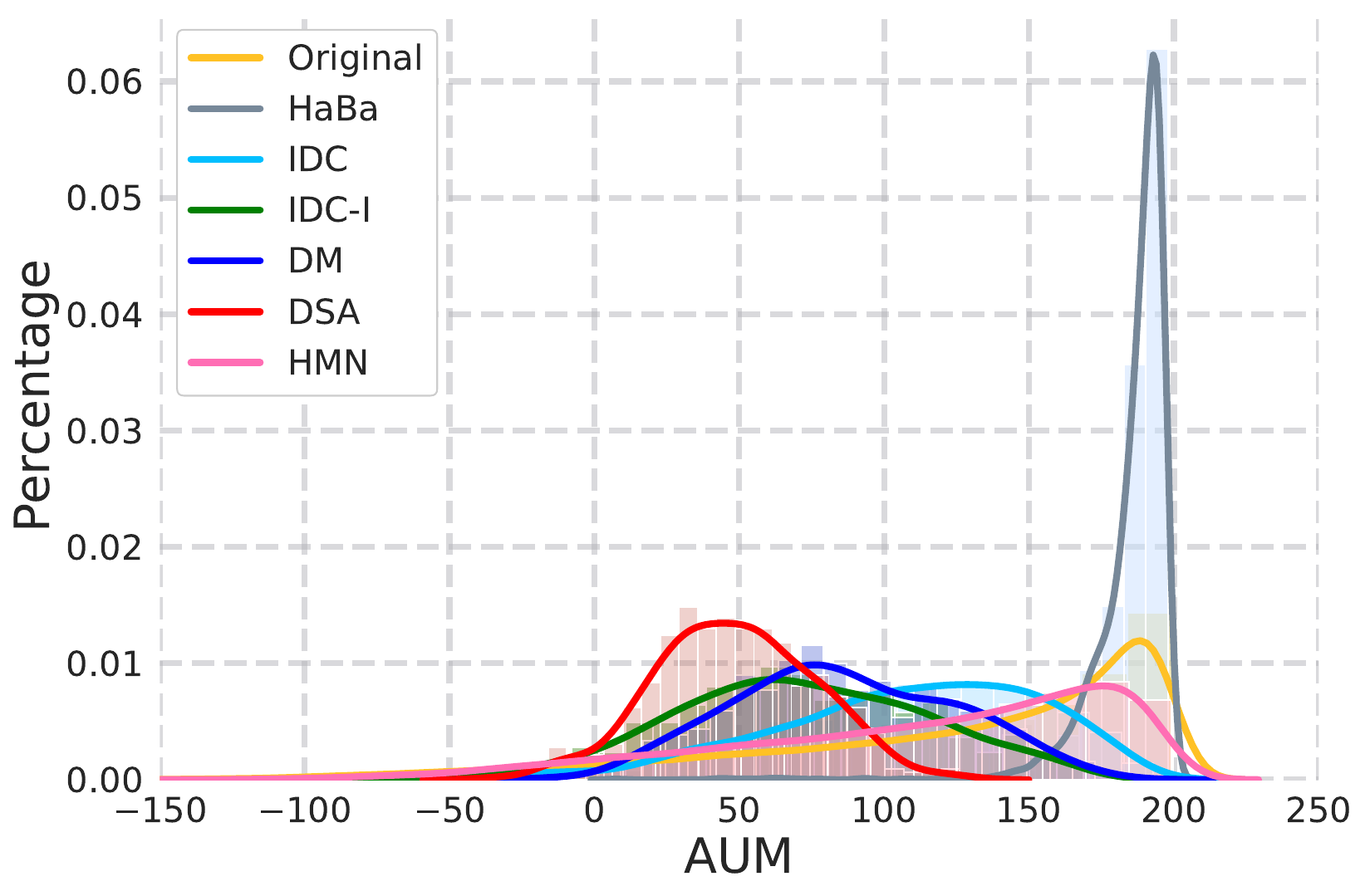}
    \caption{The distribution of AUM of CIFAR10 training images synthesized by different approaches. Different colors denote different data condensation approaches. 
    Data parameterization based methods have more redundant images.
    }
    \label{fig:app-aum-distribution}
  \end{minipage}
\end{figure}

Figure~\ref{fig:app-aum-distribution} illustrates the distribution of AUM of images synthesized by different data condensation approaches, as well as the original data, denoted as ``Original".
We calculate the AUM by training a ConvNet for 200 epochs.
We observe that approaches (IDC-I~\cite{kim2022dataset}, DM~\cite{zhao2023dataset}, and DSA~\cite{zhao2021dataset2}) that condense data into pixel space typically synthesize fewer images with a high AUM value.
In contrast, methods that rely on data parameterization, such as HaBa~\cite{liudataset}, IDC~\cite{kim2022dataset}, and HMN~\footnote{We did not evaluate LinBa due to its substantial time requirements.}, tend to produce a higher number of high-aum images. 
Notably, a large portion of images generated by HaBa exhibit an AUM value approaching 200, indicating a significant amount of redundancy that could potentially be pruned for enhanced performance.
However, due to its factorization-based design, HaBa precludes the pruning of individual images from its data containers, which limits the potential for efficiency improvements.

\begin{figure}[t]
  \centering
  \begin{minipage}{0.48\textwidth}
    \centering
    \includegraphics[width=1\textwidth]{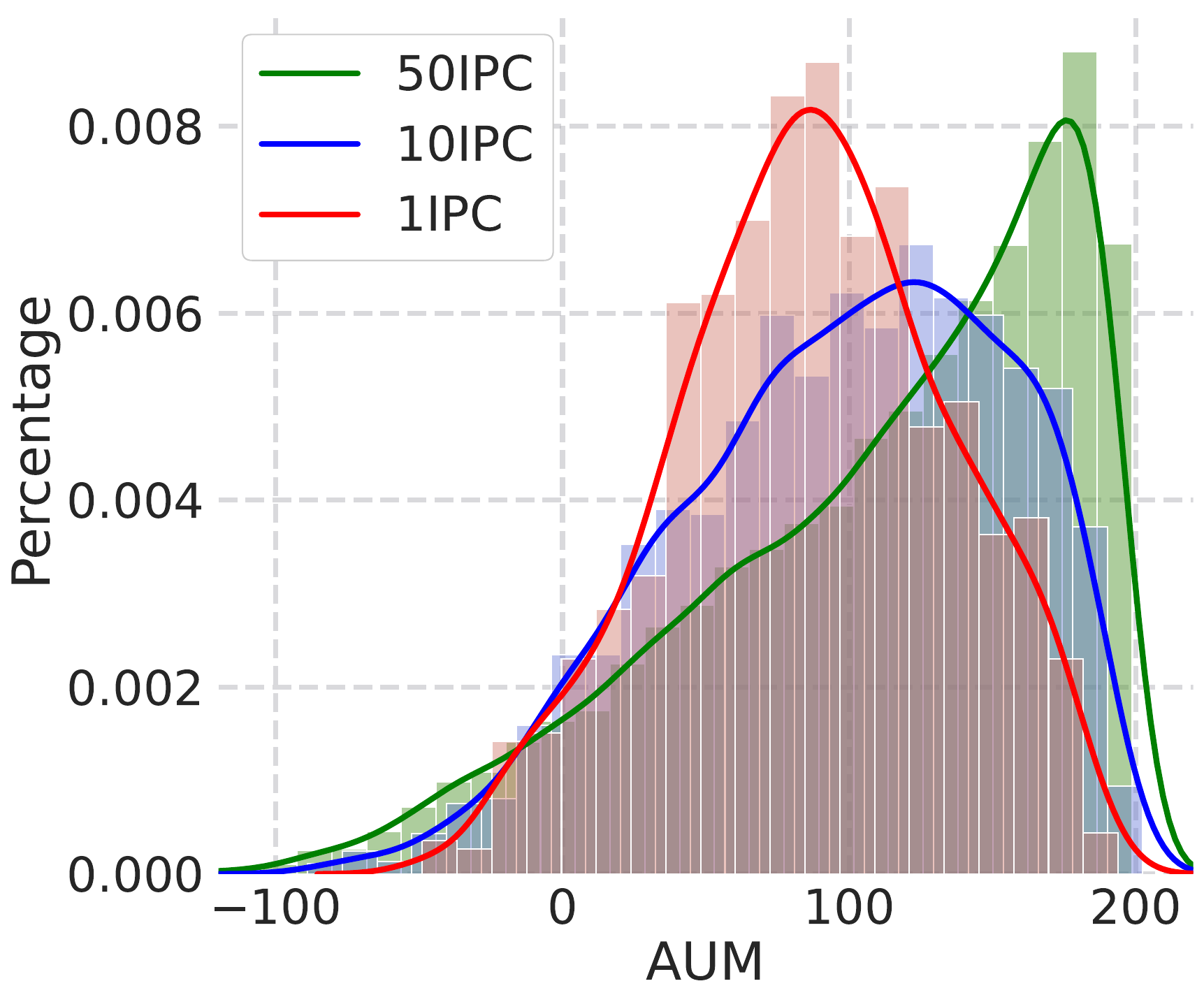}
    \caption{AUM distribution of images generated by HMNs for CIFAR10 with different storage budgets, denoted by different colors.
    }
    \label{fig:aum-distribution}
  \end{minipage}\hfill
  \begin{minipage}{0.48\textwidth}
    \centering
    \includegraphics[width=1\textwidth]{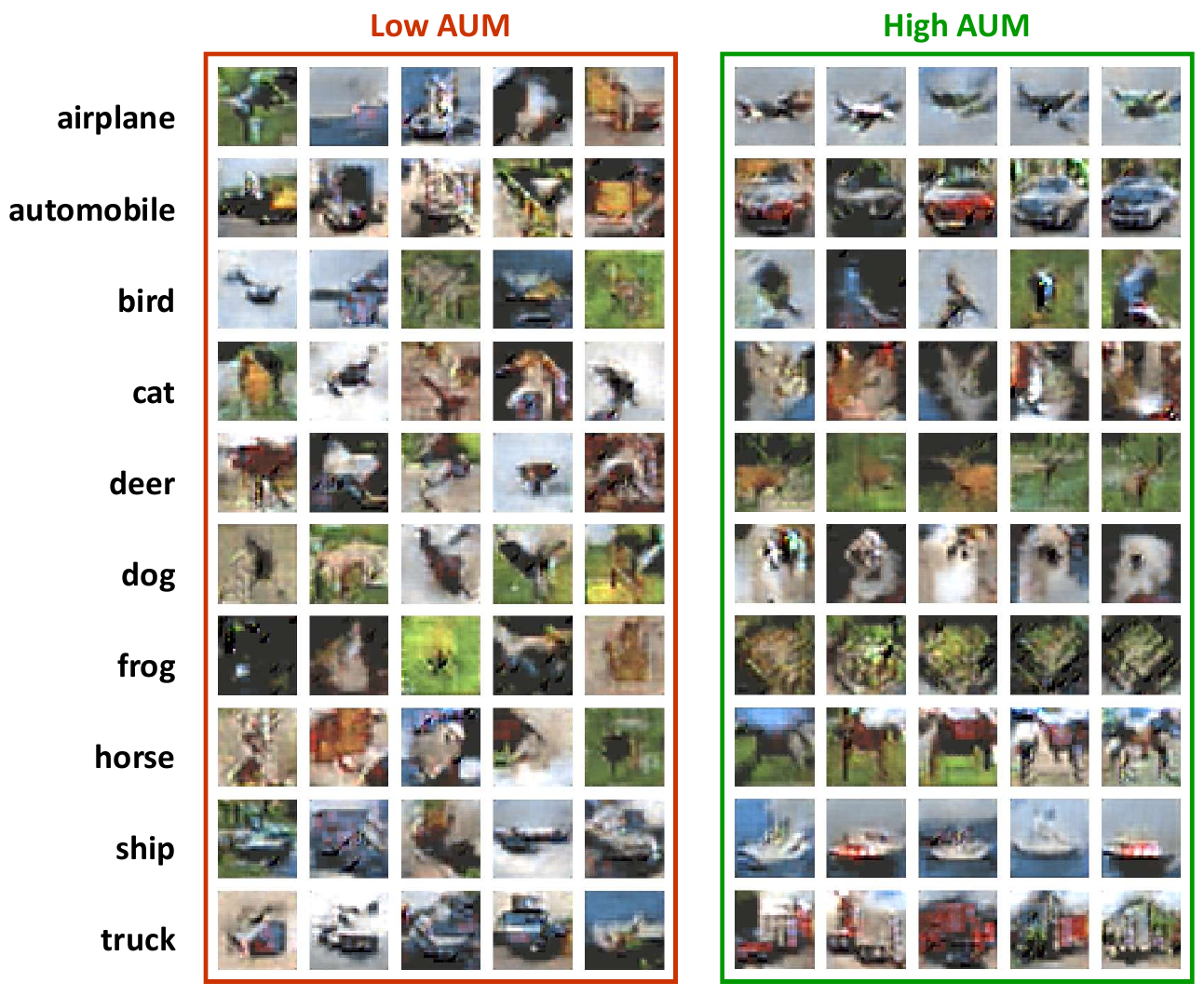}
    \caption{Visualization of the lowest and highest AUM examples generated by a 10 IPC HMN of CIFAR10. Each row represents a class.}
    \label{fig:visualization}
  \end{minipage}
\end{figure}

Moreover, we conduct a more detailed study on the images generated by HMNs.
We calculate the AUM by training a ConvNet for $200$ epochs. 
As shown in Figure~\ref{fig:aum-distribution},
many examples possess negative AUM values, indicating that they are likely hard-to-learn, low-quality images that may negatively impact training. 
Moreover, a considerable number of examples demonstrate AUM values approximating $200$, representing easy-to-learn examples that may contribute little to the training process. 
We also observe that an increased storage budget results in a higher proportion of easier examples.
This could be a potential reason why data condensation performance degrades to random selection when the storage budget keeps increasing, which is observed in \cite{cuidc}: more storage budgets add more easy examples which only provide redundant information and do not contribute much to training.
From Figure~\ref{fig:aum-distribution}, we can derive two key insights: 1) condensed datasets contain easy examples (AUM close to $200$) as well as hard examples (AUM with negative values), and 2) the proportion of easy examples varies depending on the storage budget.

Additionally, in Figure~\ref{fig:visualization}, we offer a visualization of images associated with the highest and lowest AUM values generated by an HMN. 
It is observable that images with low AUM values exhibit poor alignment with their corresponding labels, which may detrimentally impact the training process.
Conversely, images corresponding to high AUM values depict a markedly improved alignment with their classes. However, these images may be overly similar, providing limited information to training.

% \subsection{Potential Data Pruning Solution on HaBa}
% \label{ssec:app-profiling}

% \begin{wrapfigure}{r}{0.48\textwidth}
%     \vspace{-0.15cm}
%     \centering
%     \includegraphics[width=\linewidth]{figs/2dblocks.pdf}
%     \caption{Rank distribution for different basis vectors in HaBa for CIFAR10 10 IPC. Each column in this figure represents the difficulty rank of images generated using the same basis vector. The color stands for the difficulty rank among all generated images. Green denotes easy-to-learn~(less important) images, while red indicates hard-to-learn~(more important) images. }
%     % The difficulty of images generated by the same basis vector can be very different. Thus, simply eliminating a basis vector does not guarantee selective pruning of only the desired images.}
%     \label{fig:haba-basis-difficulty}
%     \vspace{-0.5cm}
% \end{wrapfigure}

% A potential solution for pruning factorization-based data containers is to prune basis vectors in the data containers (each basis vector is used to generate multiple training images).
% However, we show that directly pruning these basis vectors can lead to removing important data.
% In Figure~\ref{fig:haba-basis-difficulty}, we plot the importance rank distribution for training data generated by each basis vector.
% We observe that the difficulty of images generated by the same basis vector can differ greatly. Thus, simply pruning basis vectors does not guarantee selective pruning of only desired images.
\subsection{Visualization}
\label{ssec:app-visualization}

To provide  a better understanding of the images generated by HMNs, we visualize generated images with different AUM values on CIFAR10, CIFAR100, and SVHN with 1.1 IPC/11 IPC/55 IPC storage budgets in this section
The visualization results are presented in the following images.

Similar to what we observe in Section~3.2 in the main paper, images with a high AUM value are better aligned with their respective labels. 
Conversely, images with a low AUM value typically exhibit low image quality or inconsistencies between their content and associated labels. 
For instance, in the visualizations of SVHNs (depicted in Figures~\ref{fig:svhn-1ipc}~\ref{fig:svhn-10ipc}~\ref{fig:svhn-50ipc}),
the numbers in the generated images with a high AUM value are readily identifiable, 
but content in the generated images with a low AUM value is hard to recognize.
Those images are misaligned with their corresponding labels and can be detrimental to training.
Pruning on those images can potentially improve training performance.
Furthermore, we notice an enhancement in the quality of images generated by HMNs when more storage budgets are allocated.
This improvement could be attributable to the fact that images generated by HMNs possess an enlarged instance-level memory, as indicated in Table~\ref{tab:setting-memory}. A larger instance-level memory stores additional information, thereby contributing to better image generation quality.

From the visualization, we also find that, unlike images generated by generative models, like GAN or diffusion models, images generated by HMNs do not exhibit comparably high quality.
We would like to clarify that the goal of data condensation is not to generate high-quality images, but to generate images representing the training behavior of the original dataset. 
The training loss of data condensation can not guarantee the quality of the generated images.

% \newpage

\begin{figure}[h]
  \centering
  \begin{minipage}{0.32\textwidth}
    \centering
    \includegraphics[width=1\textwidth]{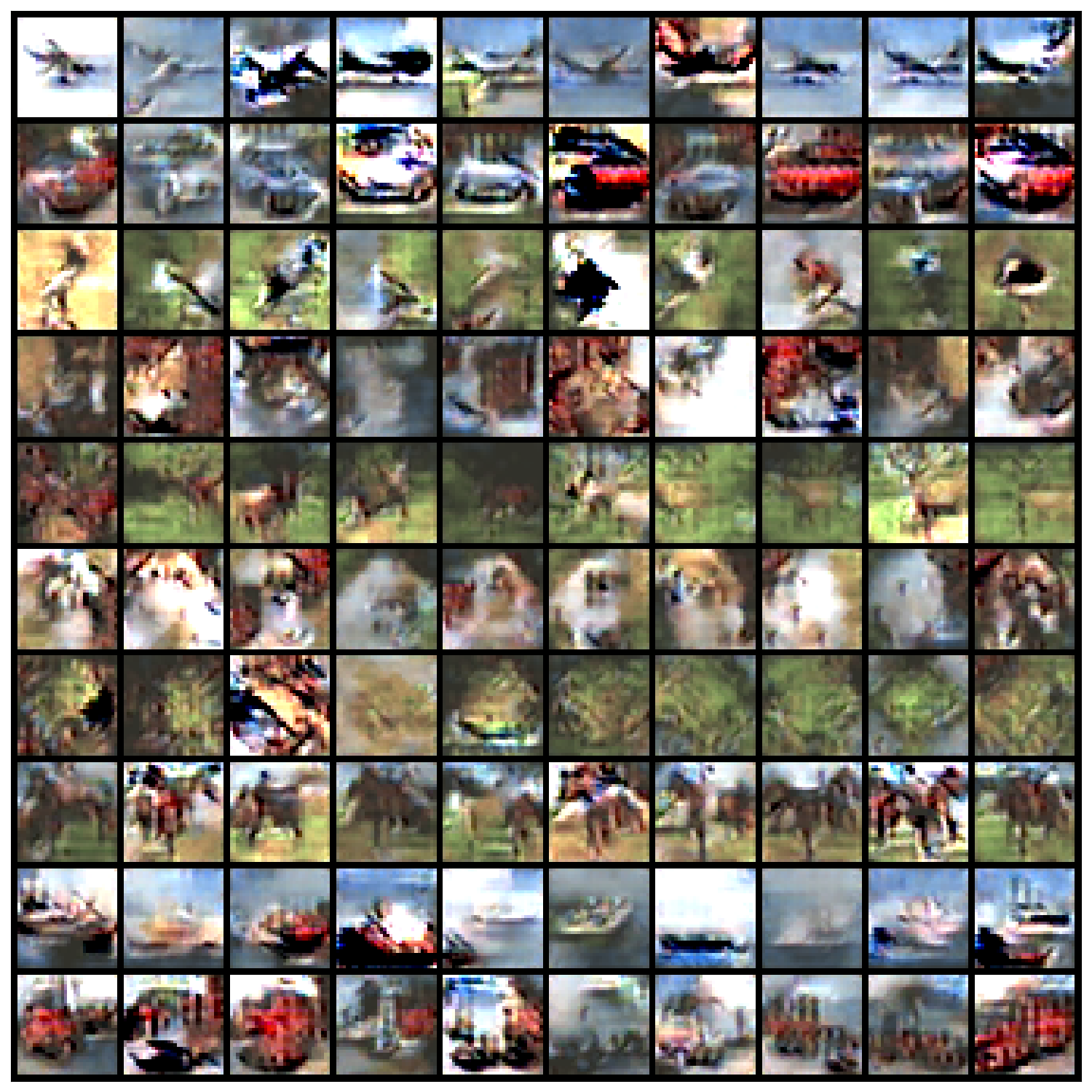}
    \centering High AUM (Easy) data
  \end{minipage}\hfill
  \begin{minipage}{0.32\textwidth}
    \centering
    \includegraphics[width=1\textwidth]{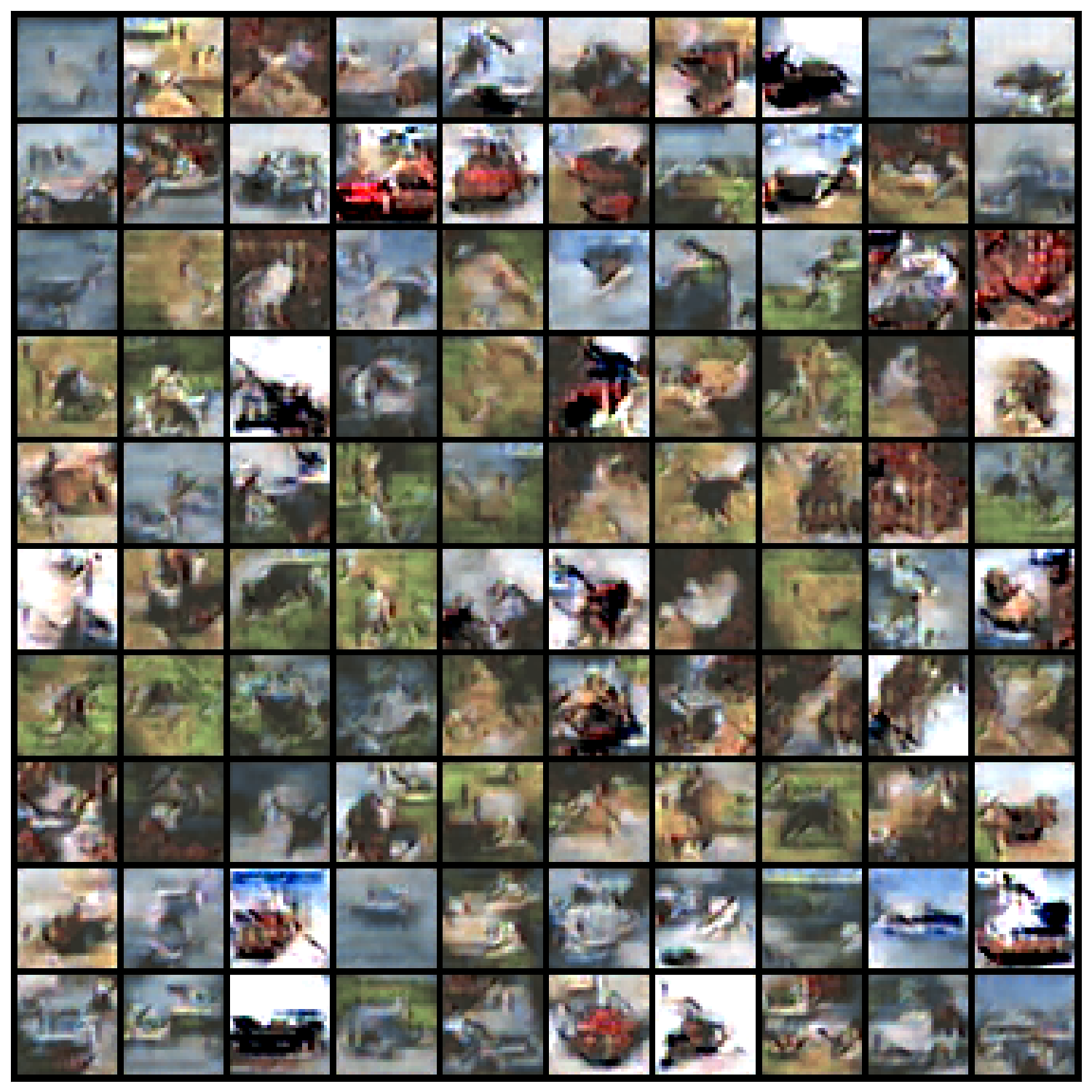}
    \centering Low AUM (Hard) data
  \end{minipage}\hfill
  \begin{minipage}{0.32\textwidth}
    \centering
    \includegraphics[width=1\textwidth]{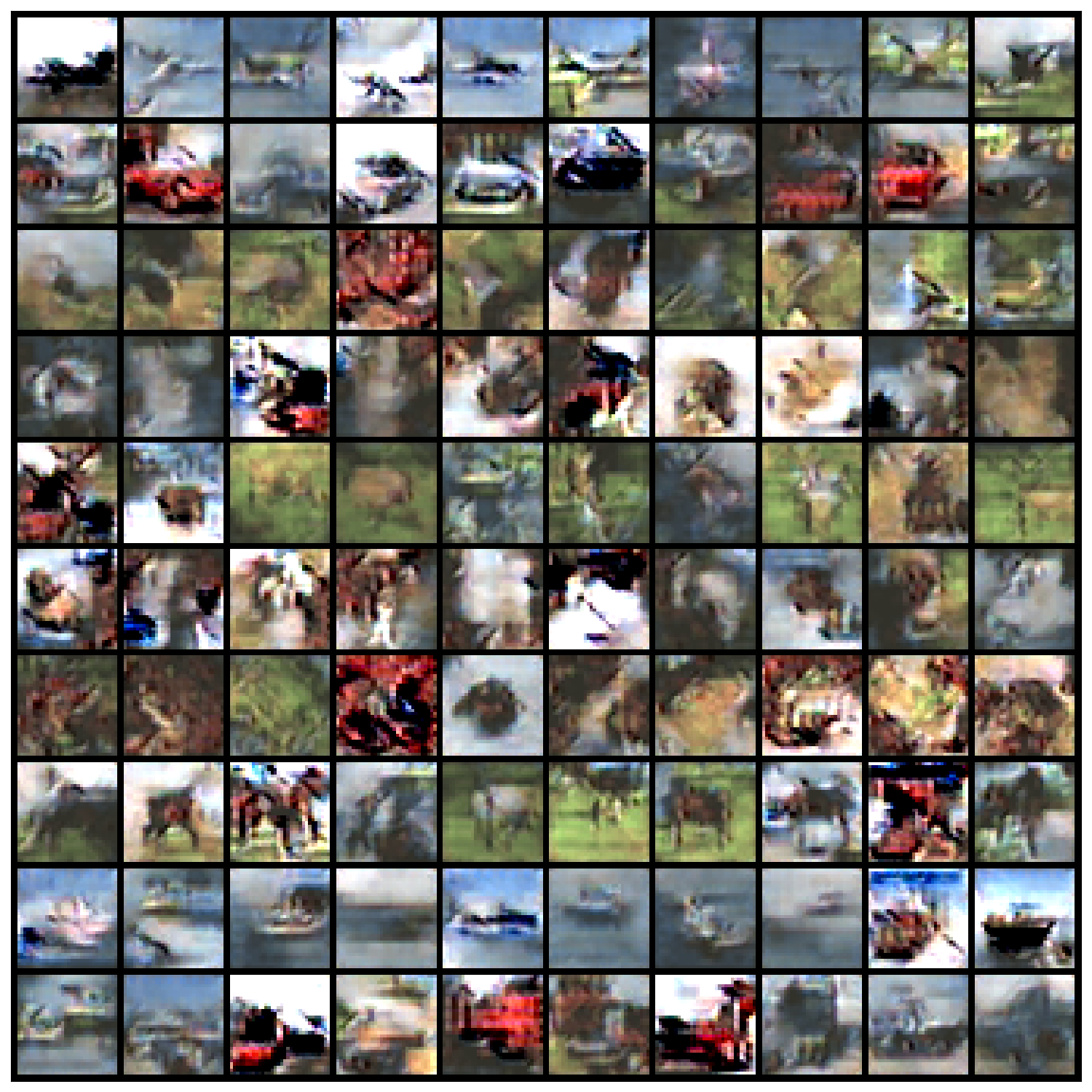}
    \centering Randomly selected data
  \end{minipage}
  \caption{Images generated by a CIFAR10 HMN with 1.1IPC storage budget. Images in each row are from the same class.}
\end{figure}

\begin{figure}[h]
  \centering
  \begin{minipage}{0.32\textwidth}
    \centering
    \includegraphics[width=1\textwidth]{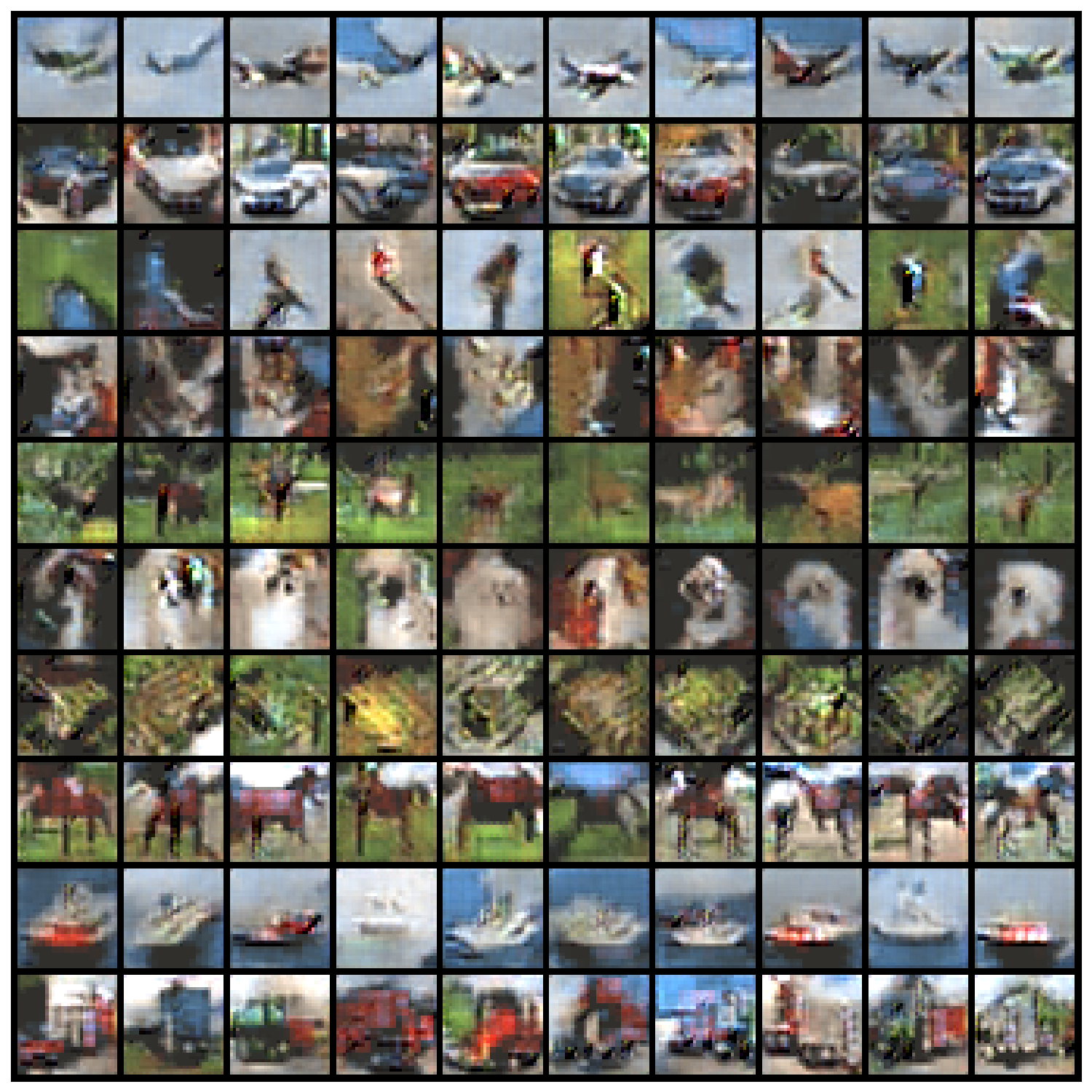}
    \centering High AUM (Easy) data
  \end{minipage}\hfill
  \begin{minipage}{0.32\textwidth}
    \centering
    \includegraphics[width=1\textwidth]{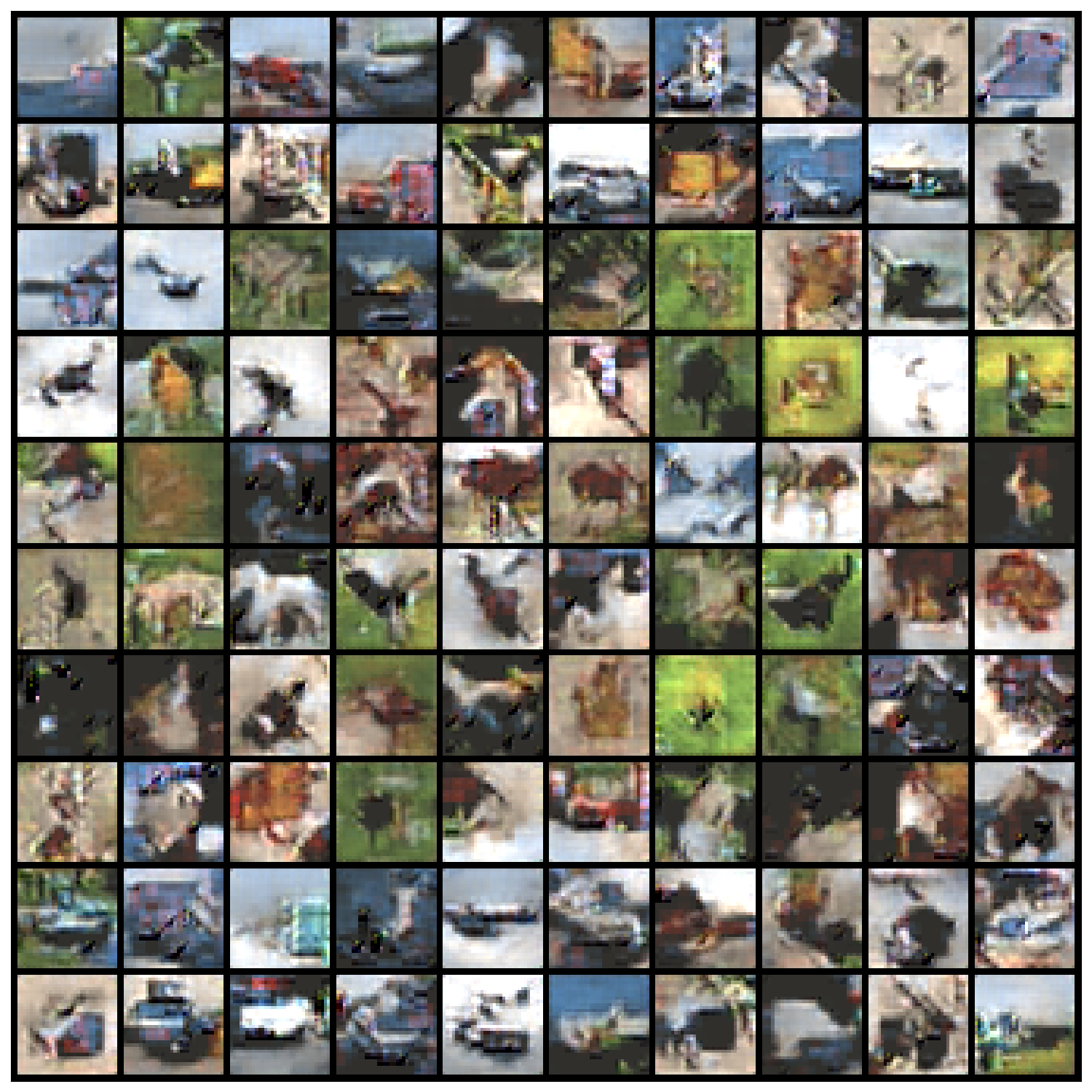}
    \centering Low AUM (Hard) data
  \end{minipage}\hfill
  \begin{minipage}{0.32\textwidth}
    \centering
    \includegraphics[width=1\textwidth]{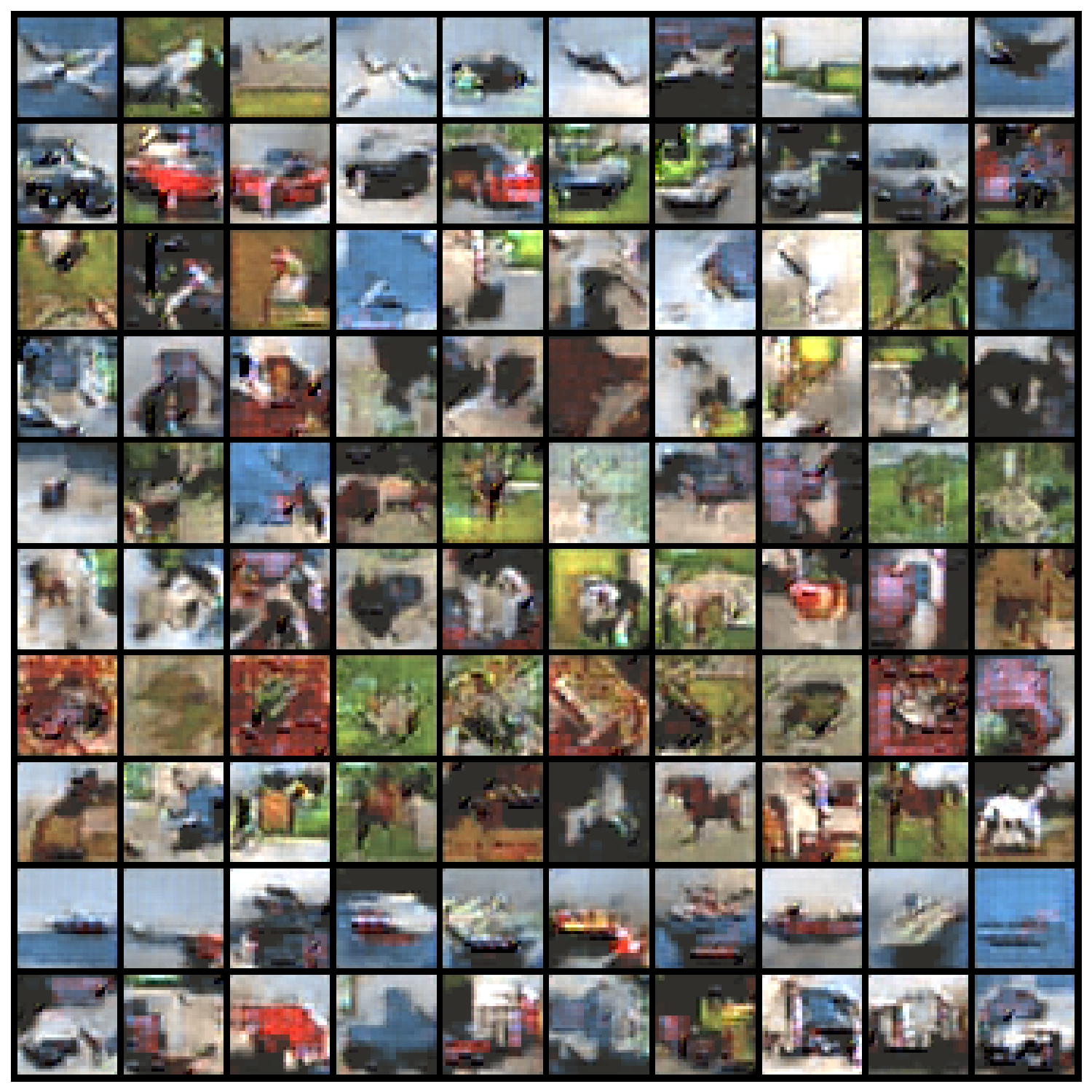}
    \centering Randomly selected data
  \end{minipage}
  \caption{Images generated by a CIFAR10 HMN with 11IPC storage budget. Images in each row are from the same class.}
\end{figure}

\begin{figure}[h]
  \centering
  \begin{minipage}{0.32\textwidth}
    \centering
    \includegraphics[width=1\textwidth]{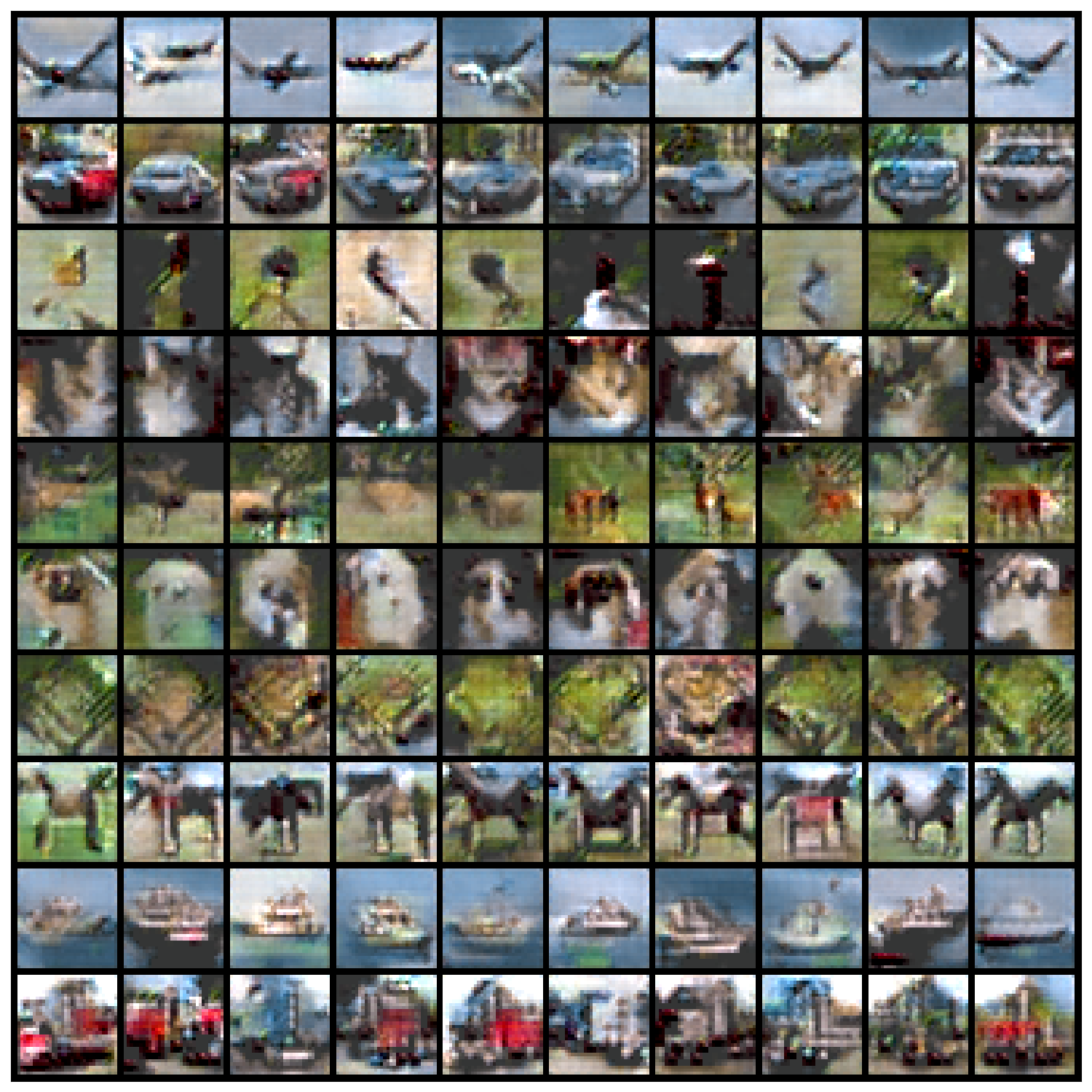}
    \centering High AUM (Easy) data
  \end{minipage}\hfill
  \begin{minipage}{0.32\textwidth}
    \centering
    \includegraphics[width=1\textwidth]{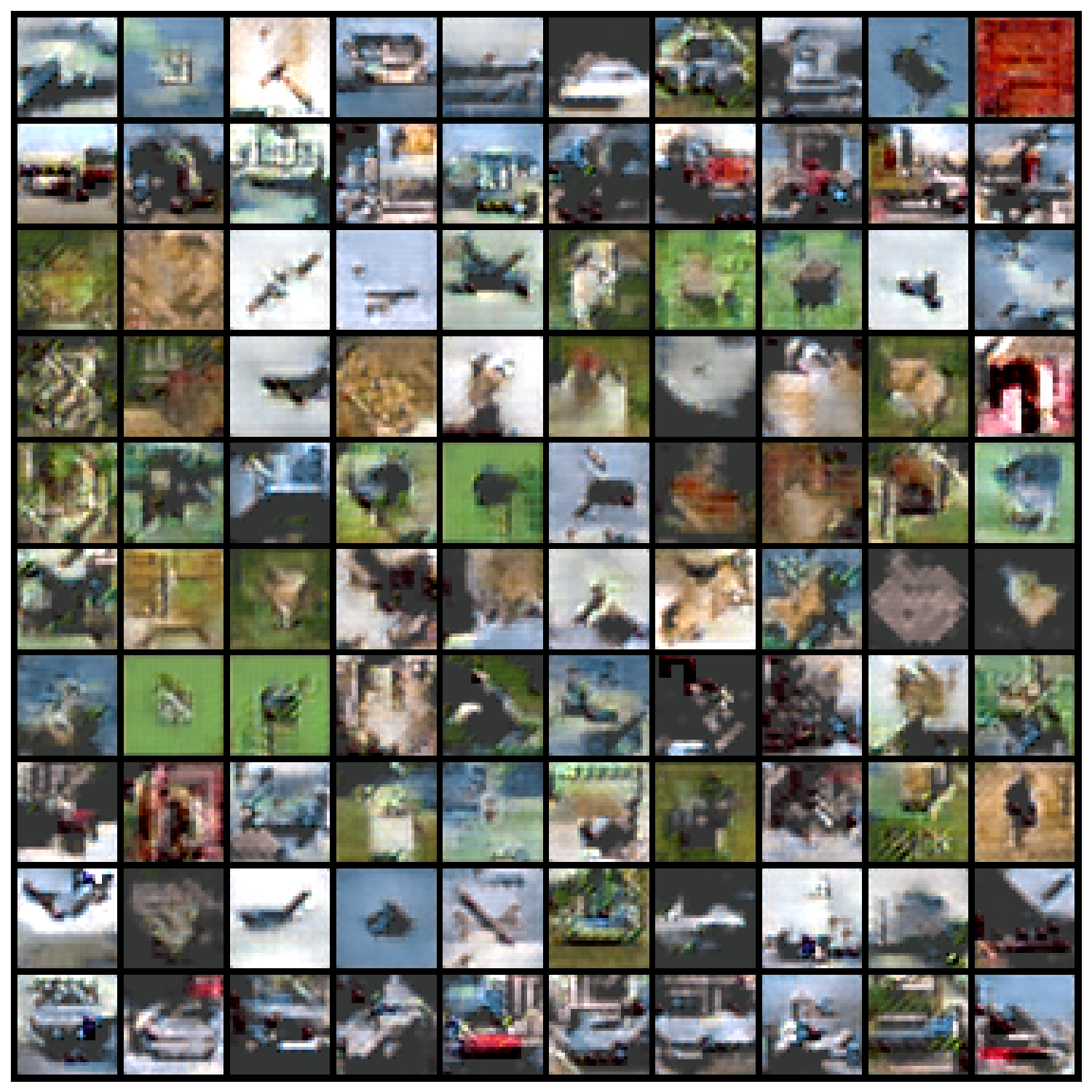}
    \centering Low AUM (Hard) data
  \end{minipage}\hfill
  \begin{minipage}{0.32\textwidth}
    \centering
    \includegraphics[width=1\textwidth]{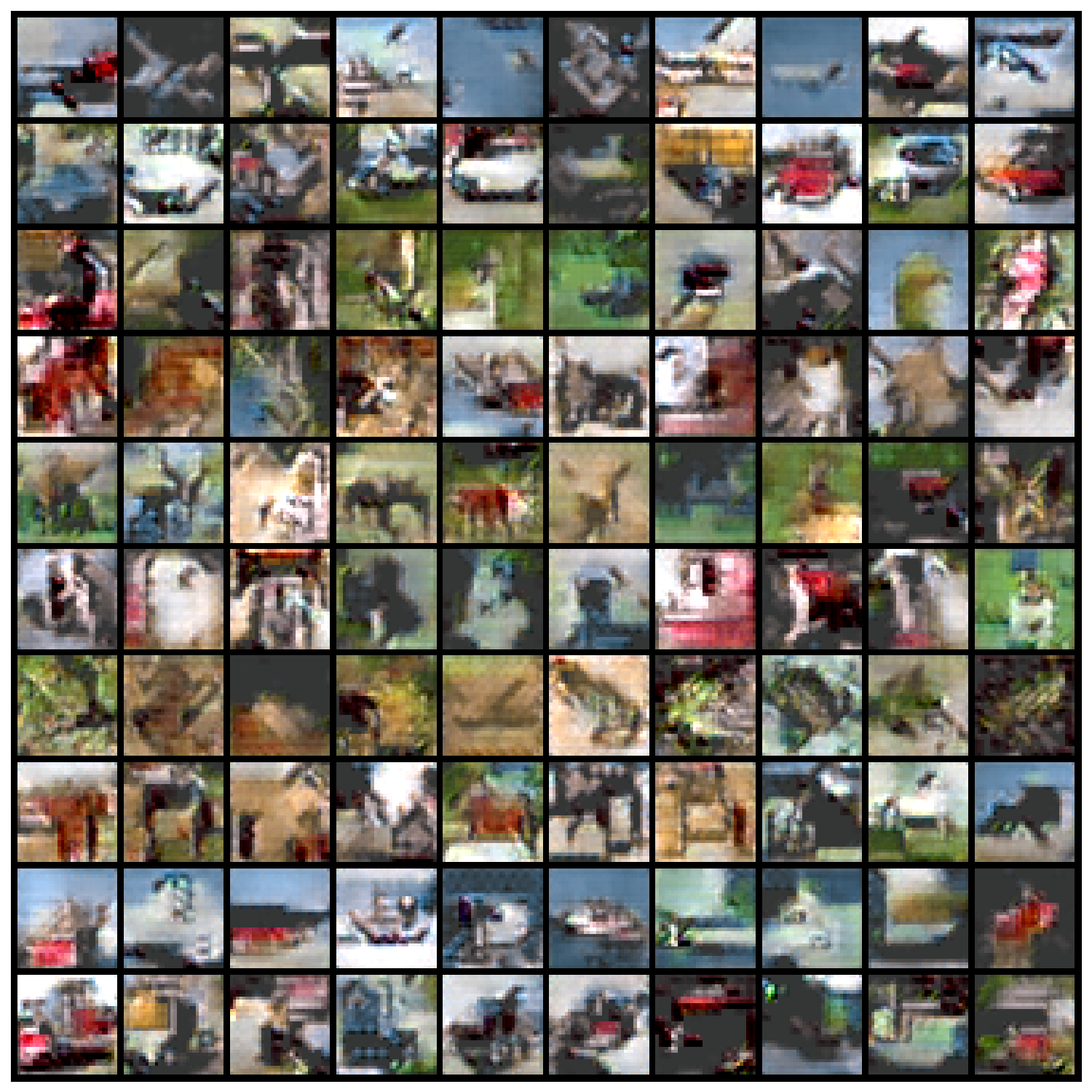}
    \centering Randomly selected data
  \end{minipage}
    \caption{Images generated by a CIFAR10 HMN with 55IPC storage budget. Images in each row are from the same class.}
\end{figure}

\begin{figure}[h]
  \centering
  \begin{minipage}{0.32\textwidth}
    \centering
    \includegraphics[width=1\textwidth]{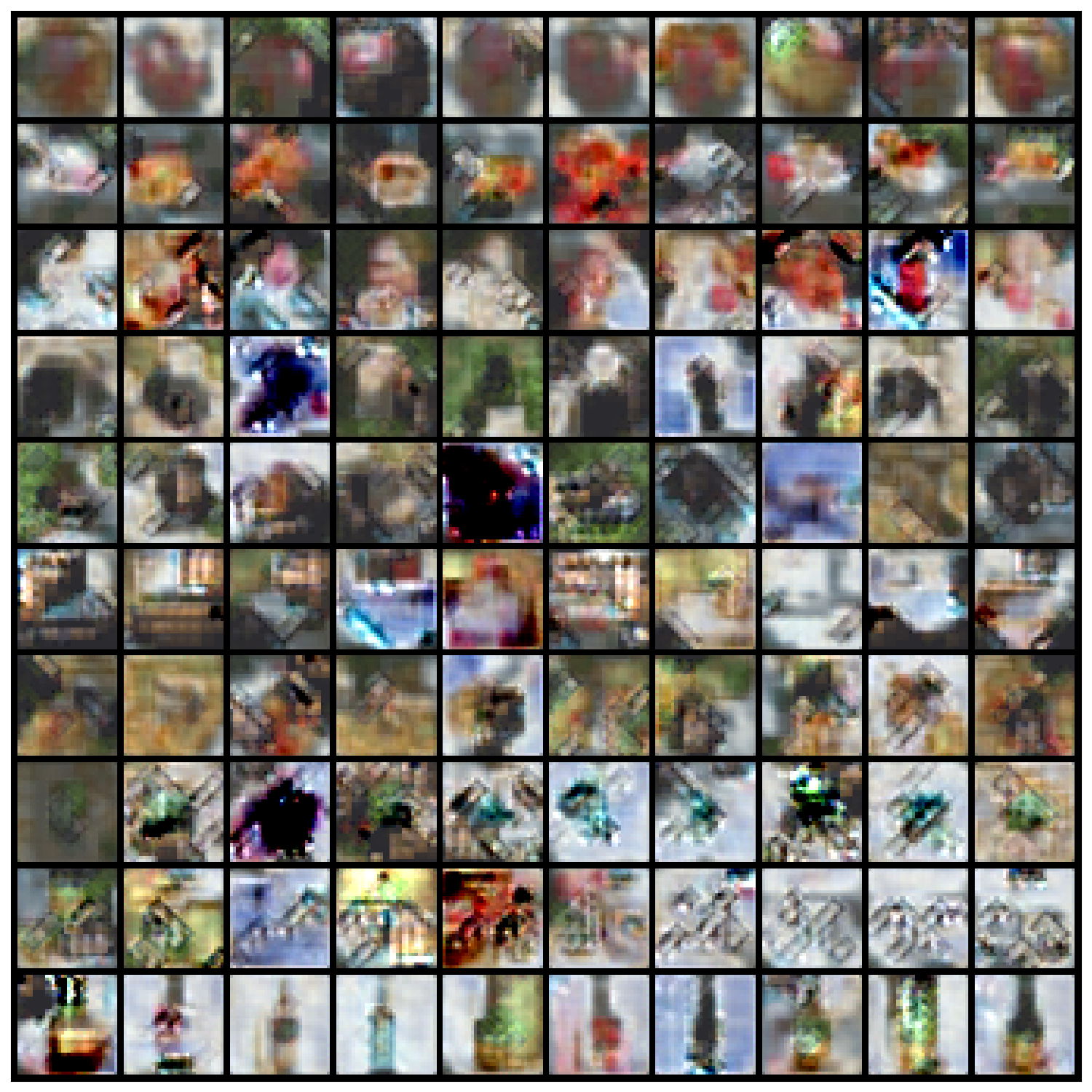}
    \centering High AUM (Easy) data
  \end{minipage}\hfill
  \begin{minipage}{0.32\textwidth}
    \centering
    \includegraphics[width=1\textwidth]{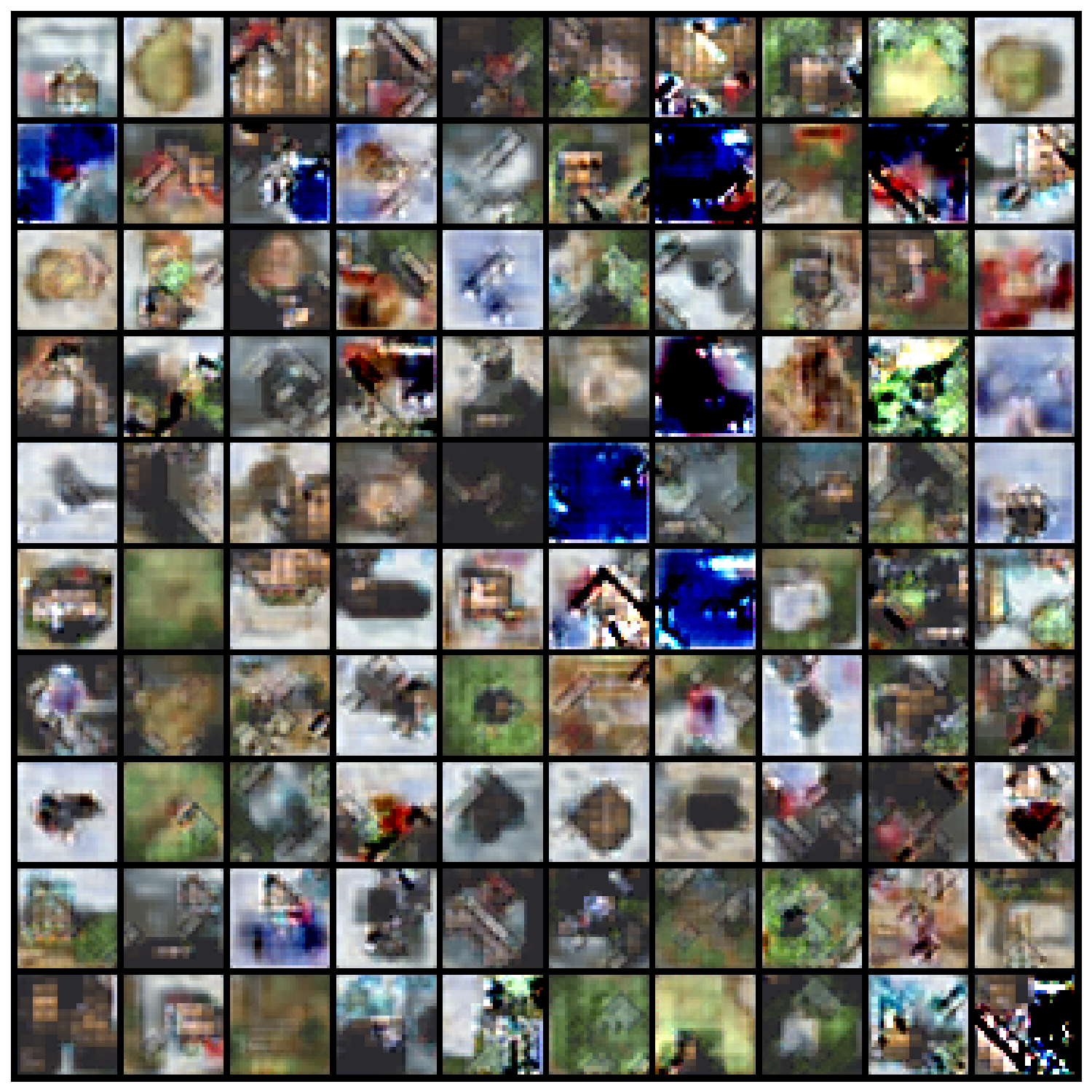}
    \centering Low AUM (Hard) data
  \end{minipage}\hfill
  \begin{minipage}{0.32\textwidth}
    \centering
    \includegraphics[width=1\textwidth]{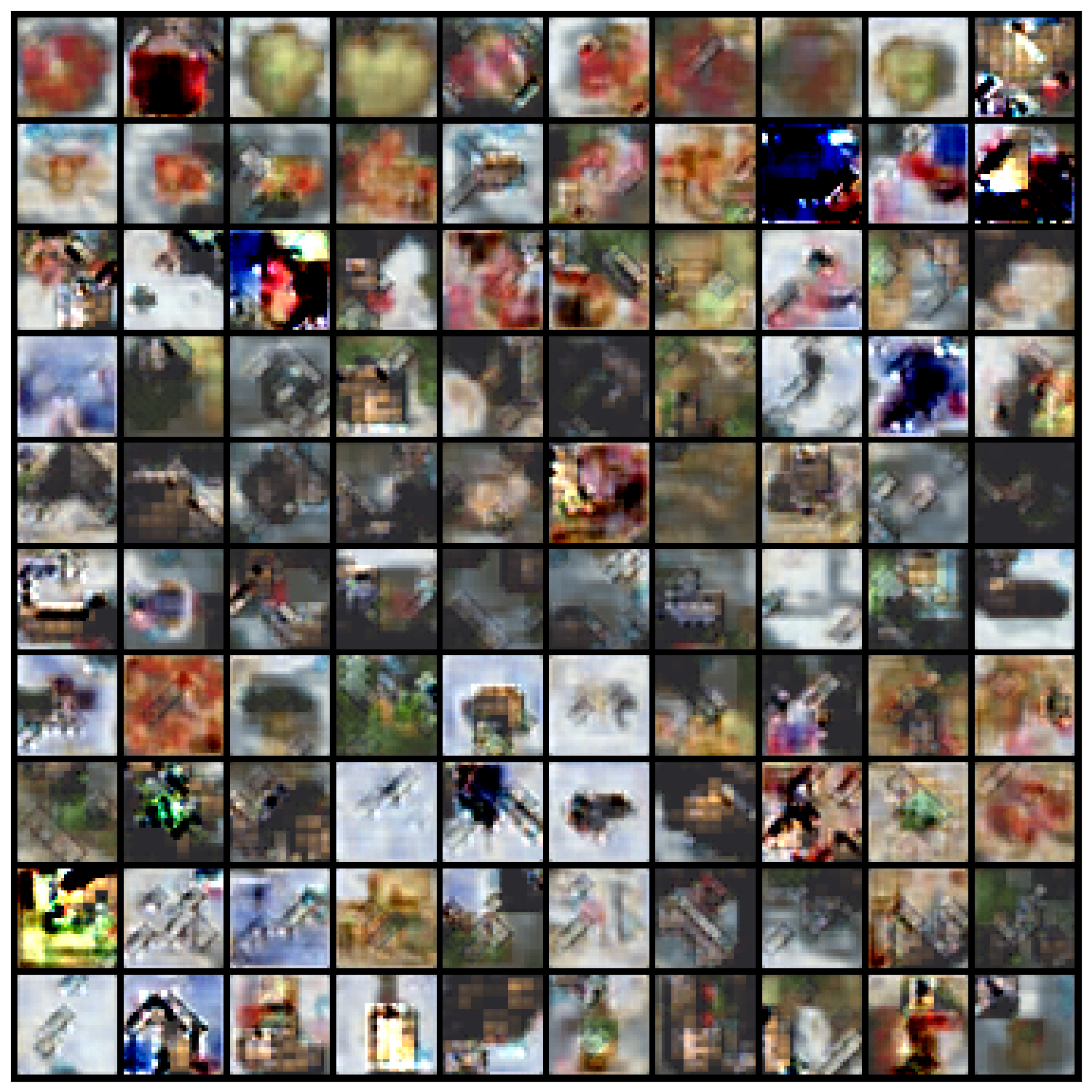}
    \centering Randomly selected data
  \end{minipage}
  \caption{Images generated by a CIFAR100 HMN with 1.1IPC storage budget. Images in each row are from the same class. We only visualize 10 classes with the smallest class number in the dataset.}
\end{figure}

\begin{figure}[h]
  \centering
  \begin{minipage}{0.32\textwidth}
    \centering
    \includegraphics[width=1\textwidth]{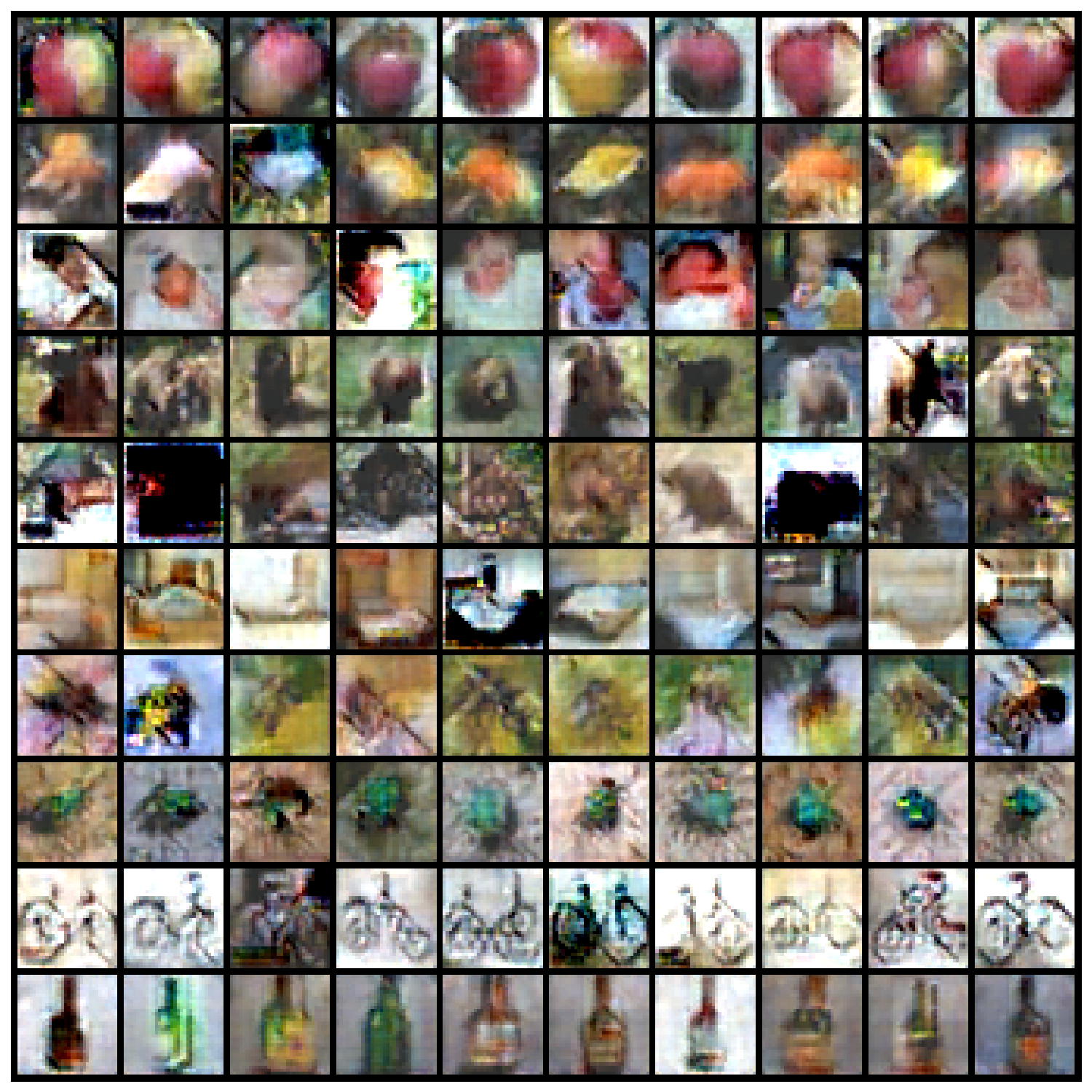}
    \centering High AUM (Easy) data
  \end{minipage}\hfill
  \begin{minipage}{0.32\textwidth}
    \centering
    \includegraphics[width=1\textwidth]{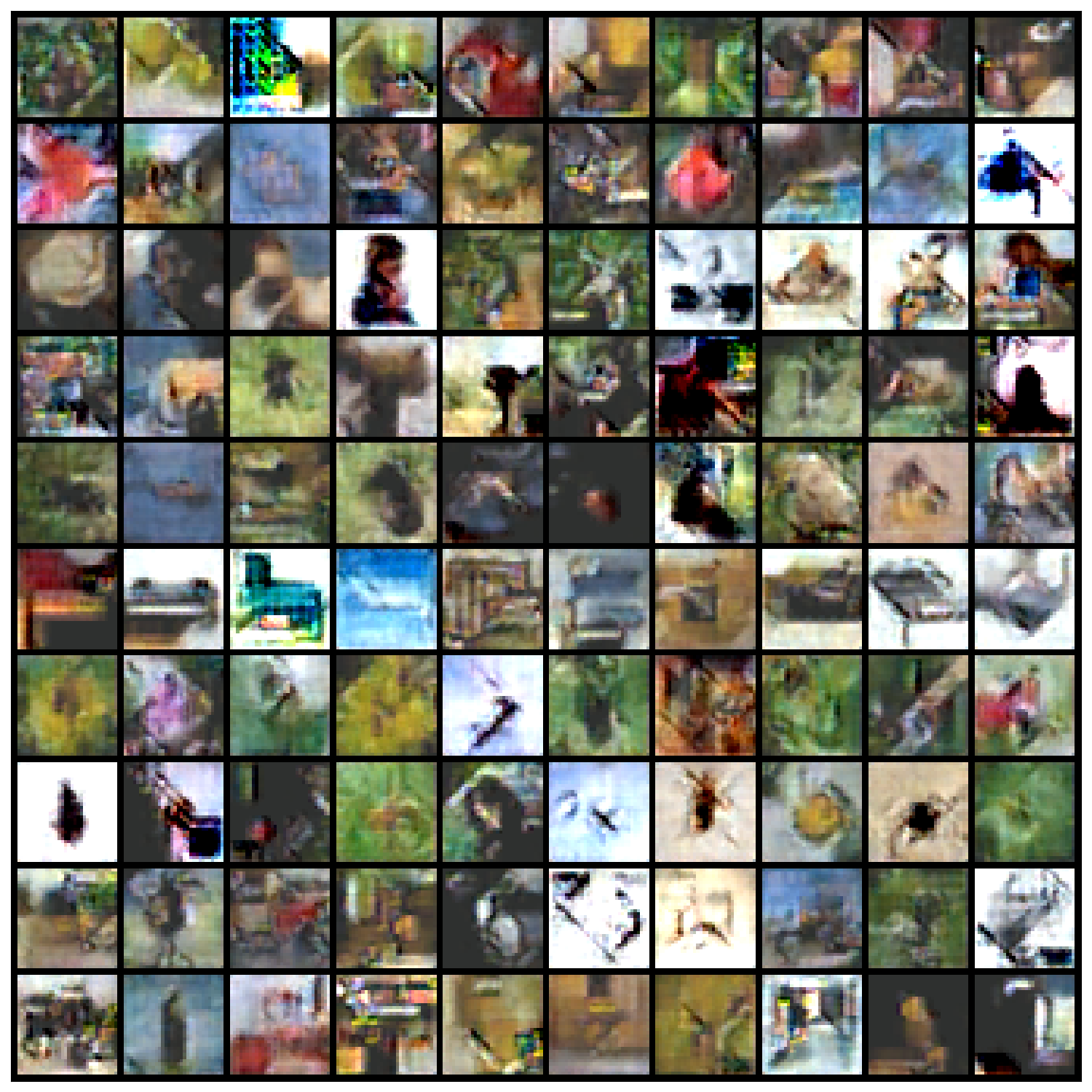}
    \centering Low AUM (Hard) data
  \end{minipage}\hfill
  \begin{minipage}{0.32\textwidth}
    \centering
    \includegraphics[width=1\textwidth]{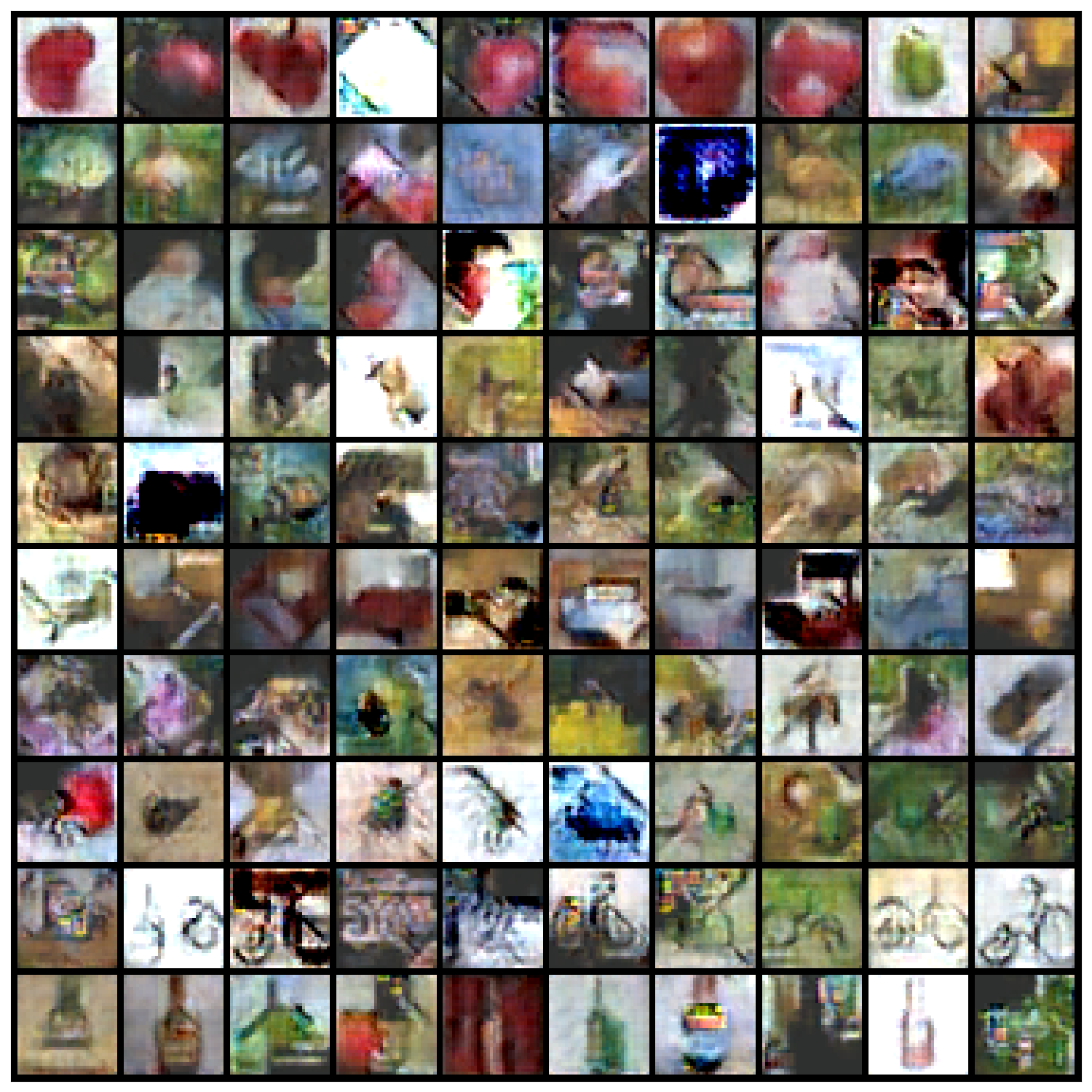}
    \centering Randomly selected data
  \end{minipage}
  \caption{Images generated by a CIFAR100 HMN with 11IPC storage budget. Images in each row are from the same class. We only visualize 10 classes with the smallest class number in the dataset.}
\end{figure}

\begin{figure}[h]
  \centering
  \begin{minipage}{0.32\textwidth}
    \centering
    \includegraphics[width=1\textwidth]{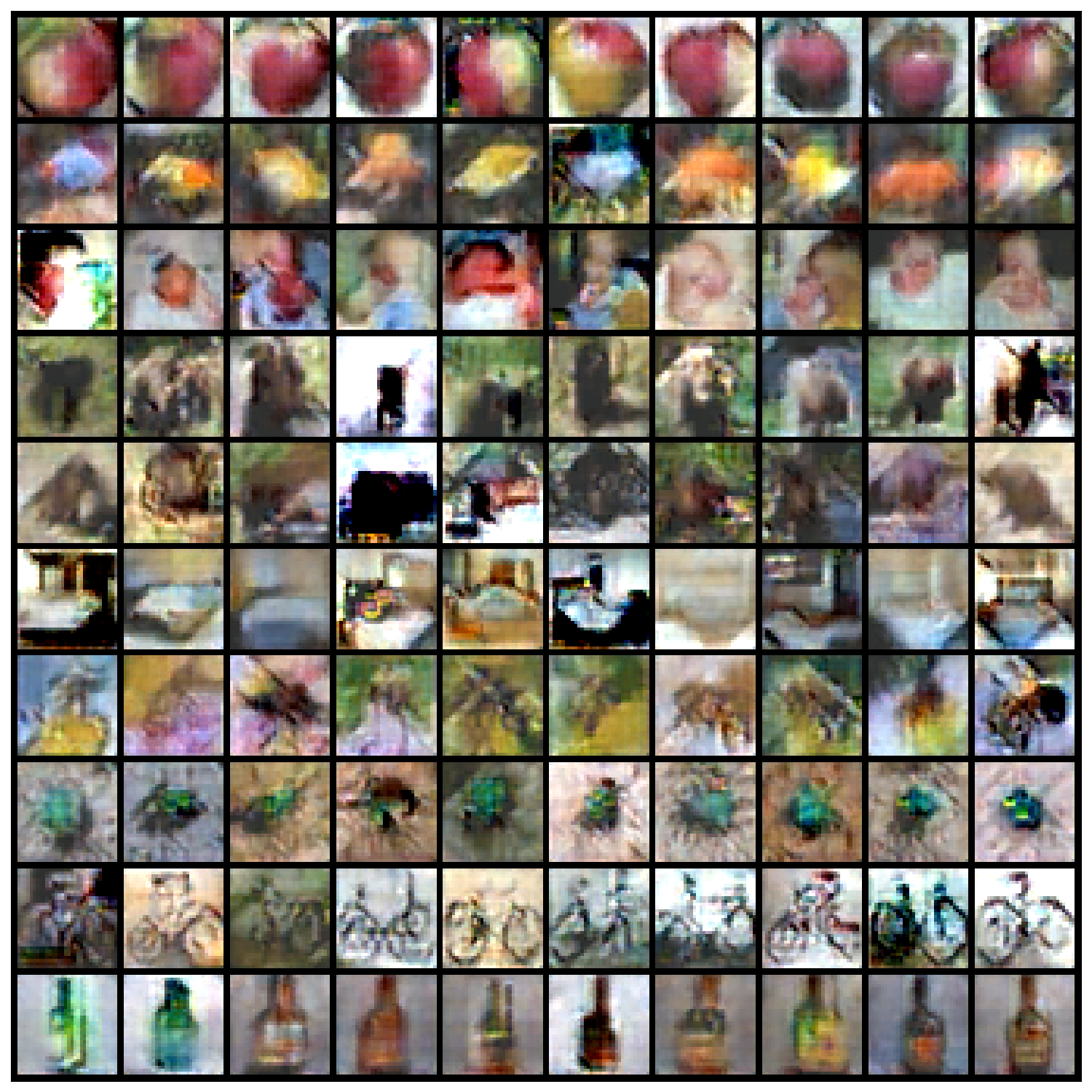}
    \centering High AUM (Easy) data
  \end{minipage}\hfill
  \begin{minipage}{0.32\textwidth}
    \centering
    \includegraphics[width=1\textwidth]{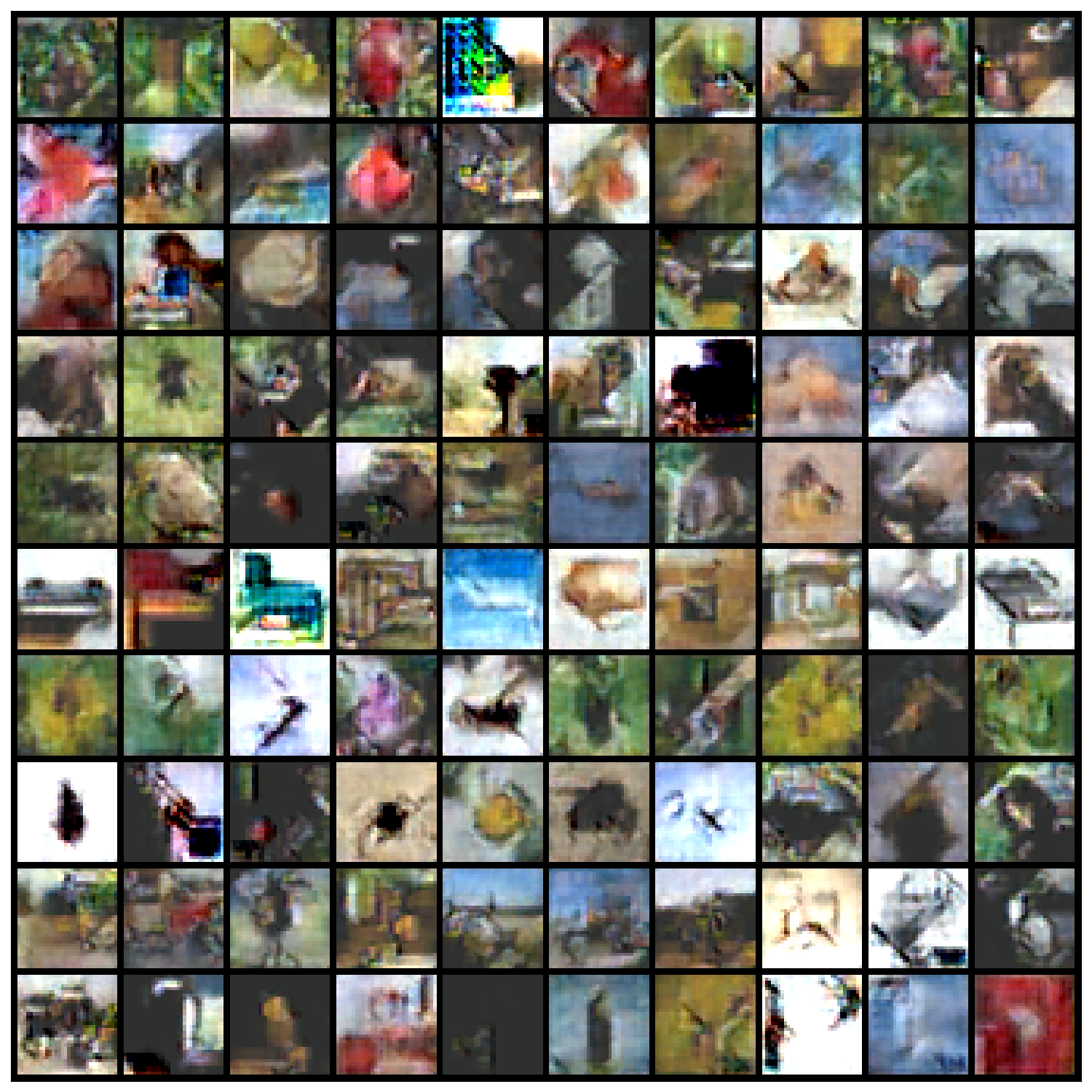}
    \centering Low AUM (Hard) data
  \end{minipage}\hfill
  \begin{minipage}{0.32\textwidth}
    \centering
    \includegraphics[width=1\textwidth]{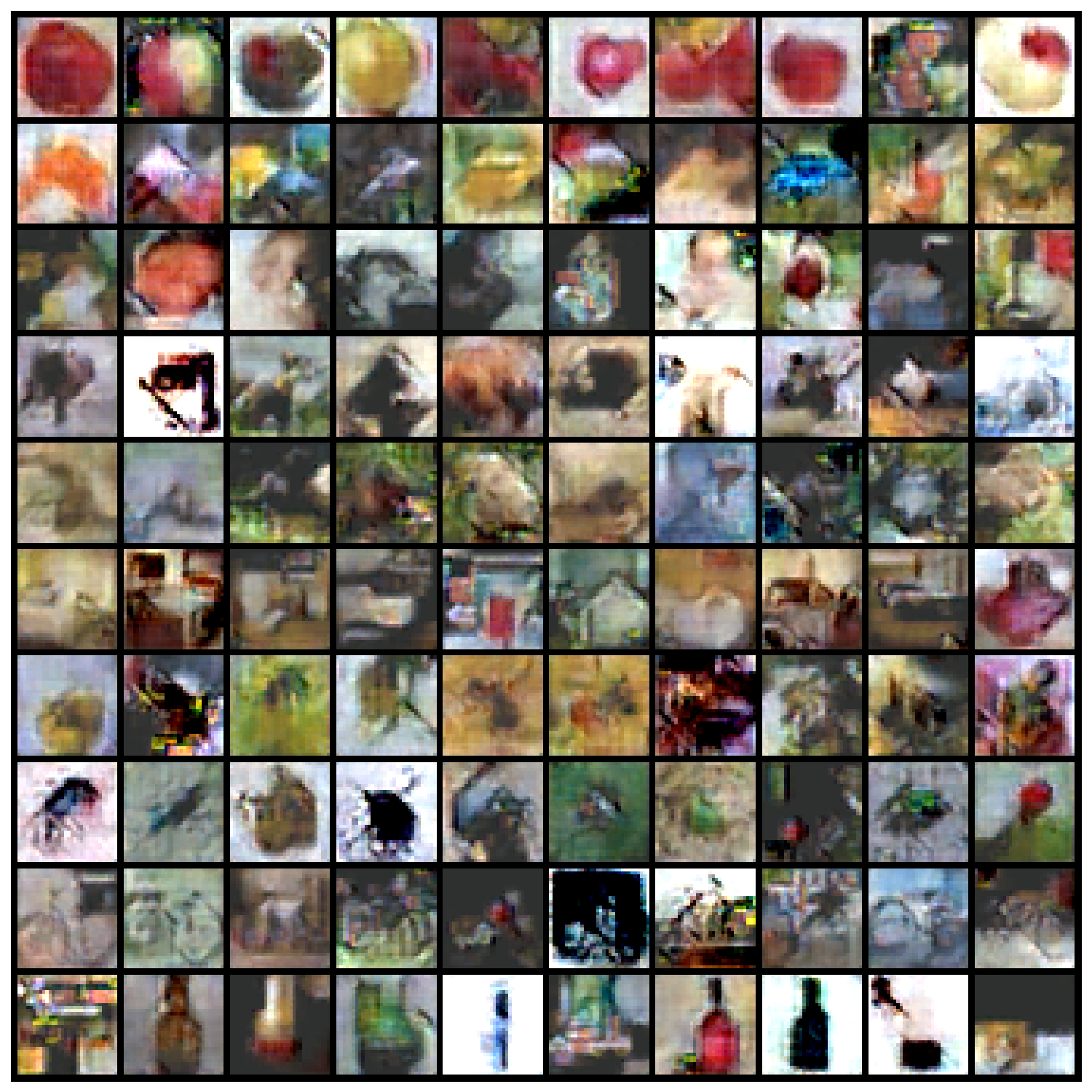}
    \centering Randomly selected data
  \end{minipage}
  \caption{Images generated by a CIFAR100 HMN with 55IPC storage budget. Images in each row are from the same class. We only visualize 10 classes with the smallest class number in the dataset.}
\end{figure}

\begin{figure}[h]
  \centering
  \begin{minipage}{0.32\textwidth}
    \centering
    \includegraphics[width=1\textwidth]{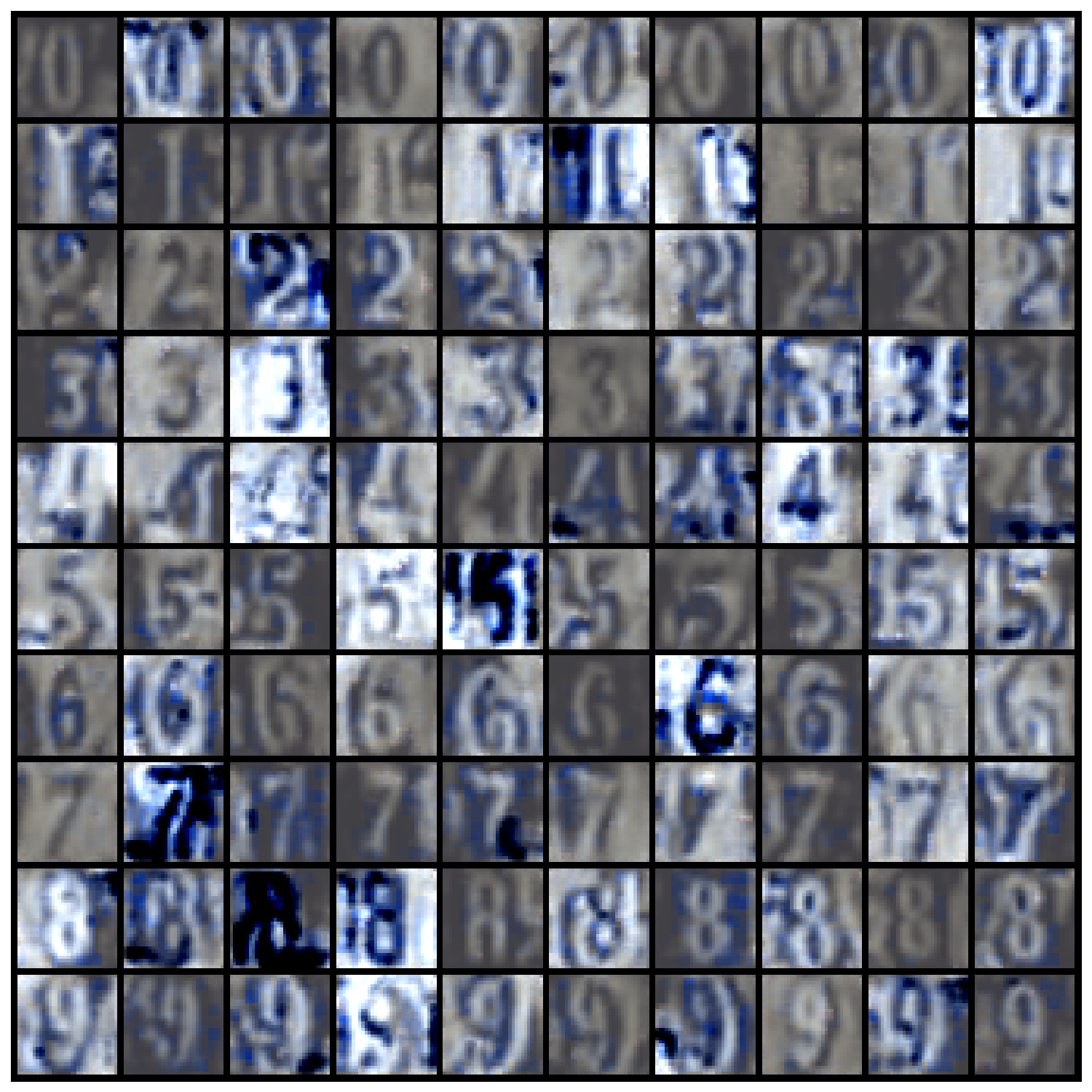}
    \centering High AUM (Easy) data
  \end{minipage}\hfill
  \begin{minipage}{0.32\textwidth}
    \centering
    \includegraphics[width=1\textwidth]{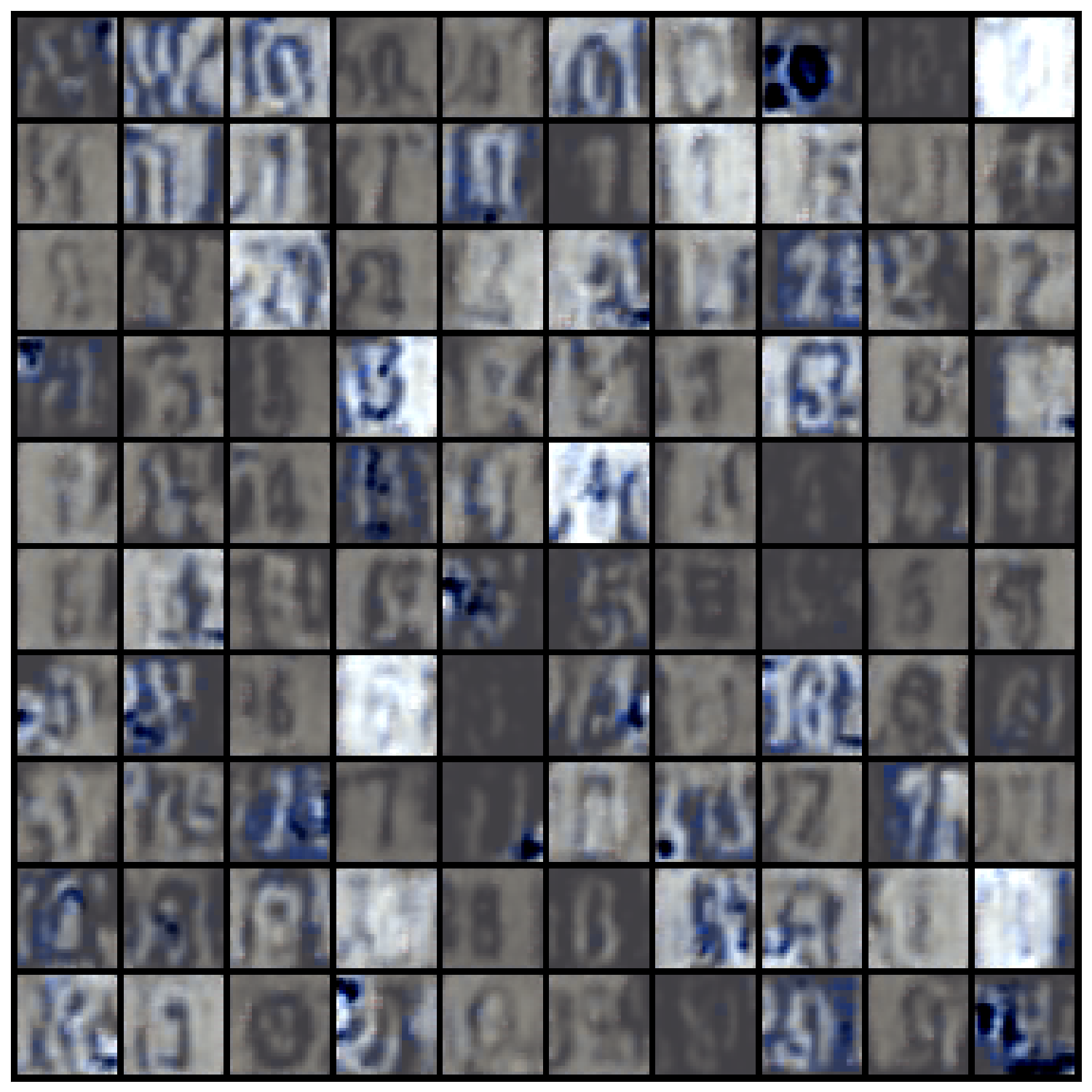}
    \centering Low AUM (Hard) data
  \end{minipage}\hfill
  \begin{minipage}{0.32\textwidth}
    \centering
    \includegraphics[width=1\textwidth]{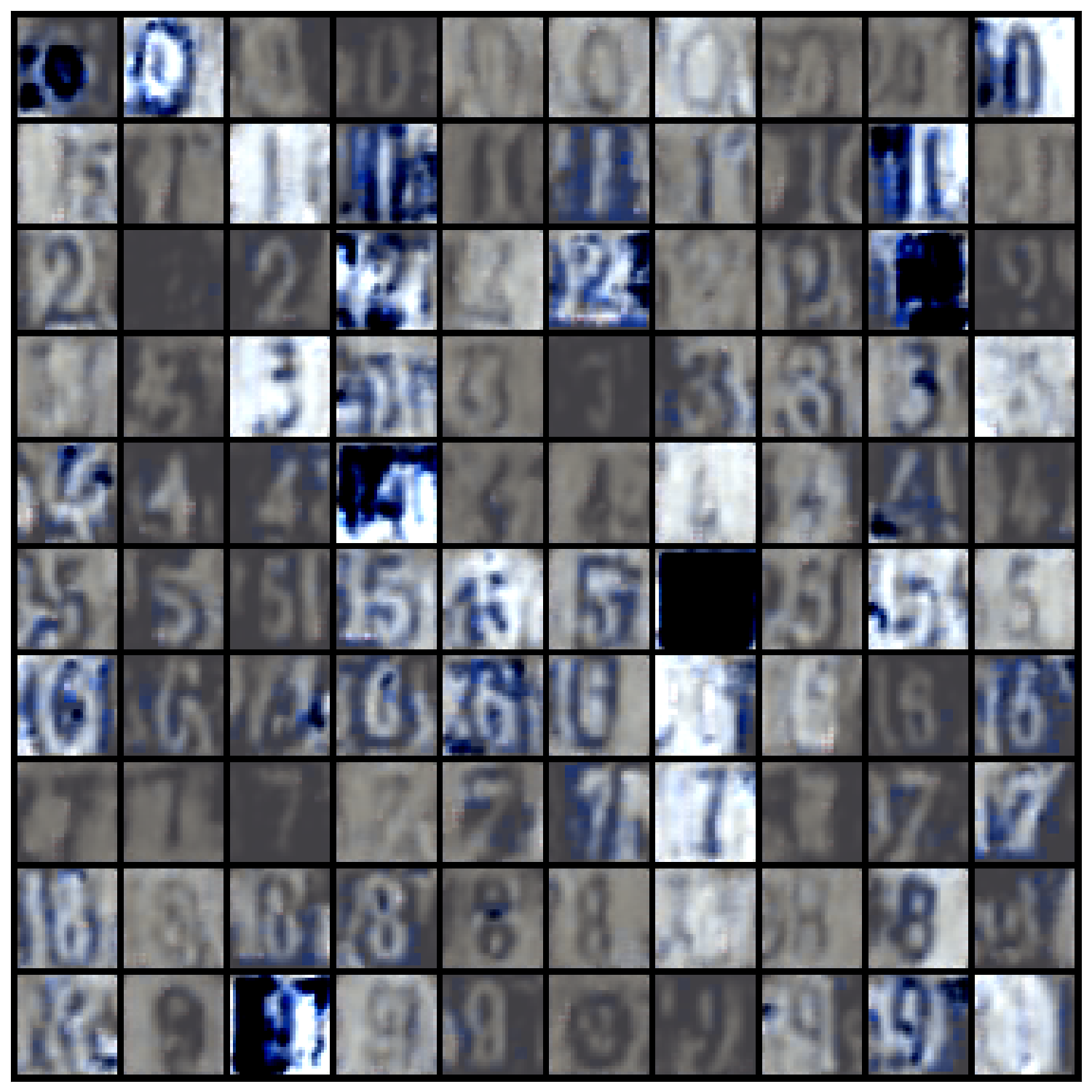}
    \centering Randomly selected data
  \end{minipage}
  \caption{Images generated by an SVHN HMN with 1.1IPC storage budget. Images in each row are from the same class. Images with a low aum value are not well-aligned with its label and can be harmful for the training.}
  \label{fig:svhn-1ipc}
\end{figure}

\begin{figure}[h]
  \centering
  \begin{minipage}{0.32\textwidth}
    \centering
    \includegraphics[width=1\textwidth]{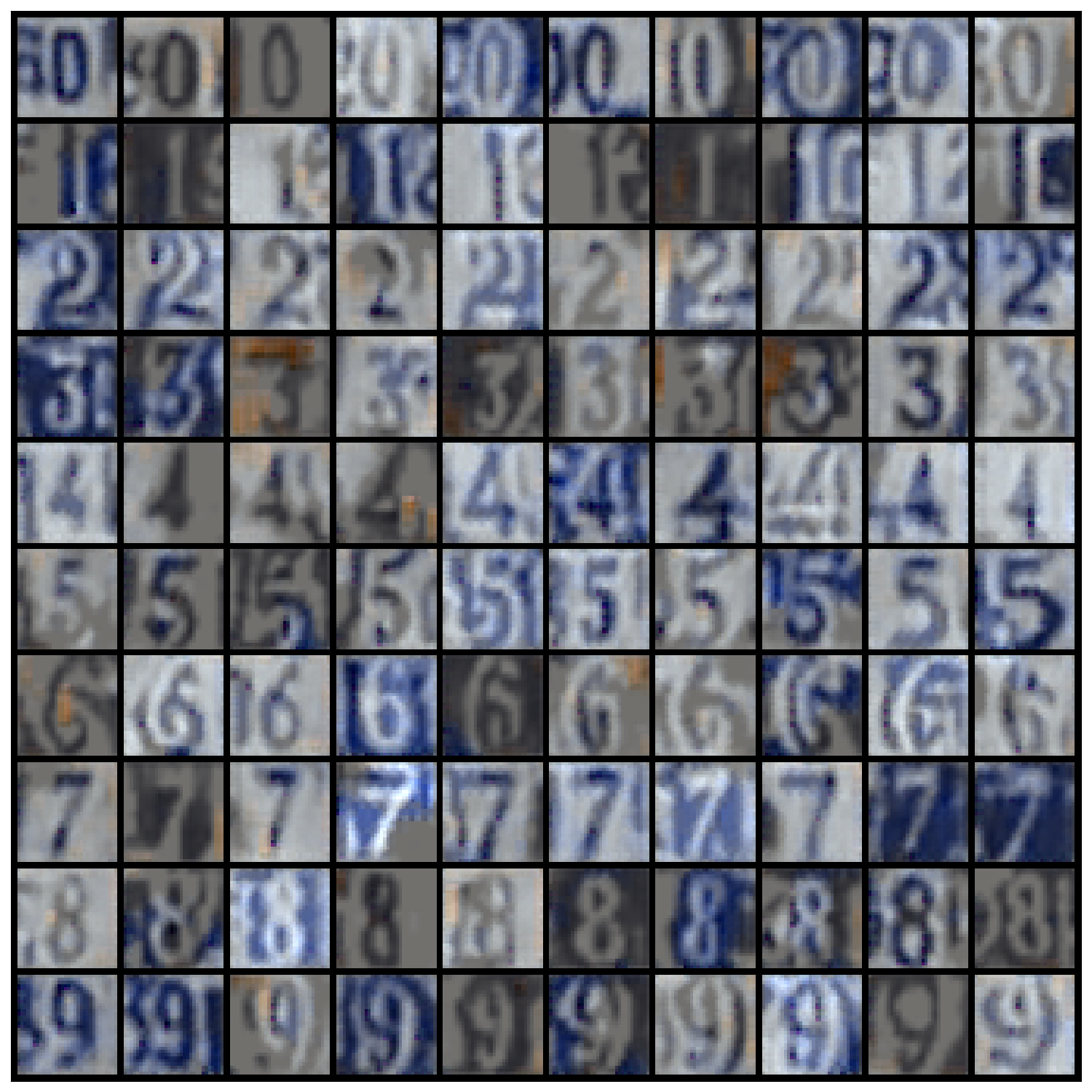}
    \centering High AUM (Easy) data
  \end{minipage}\hfill
  \begin{minipage}{0.32\textwidth}
    \centering
    \includegraphics[width=1\textwidth]{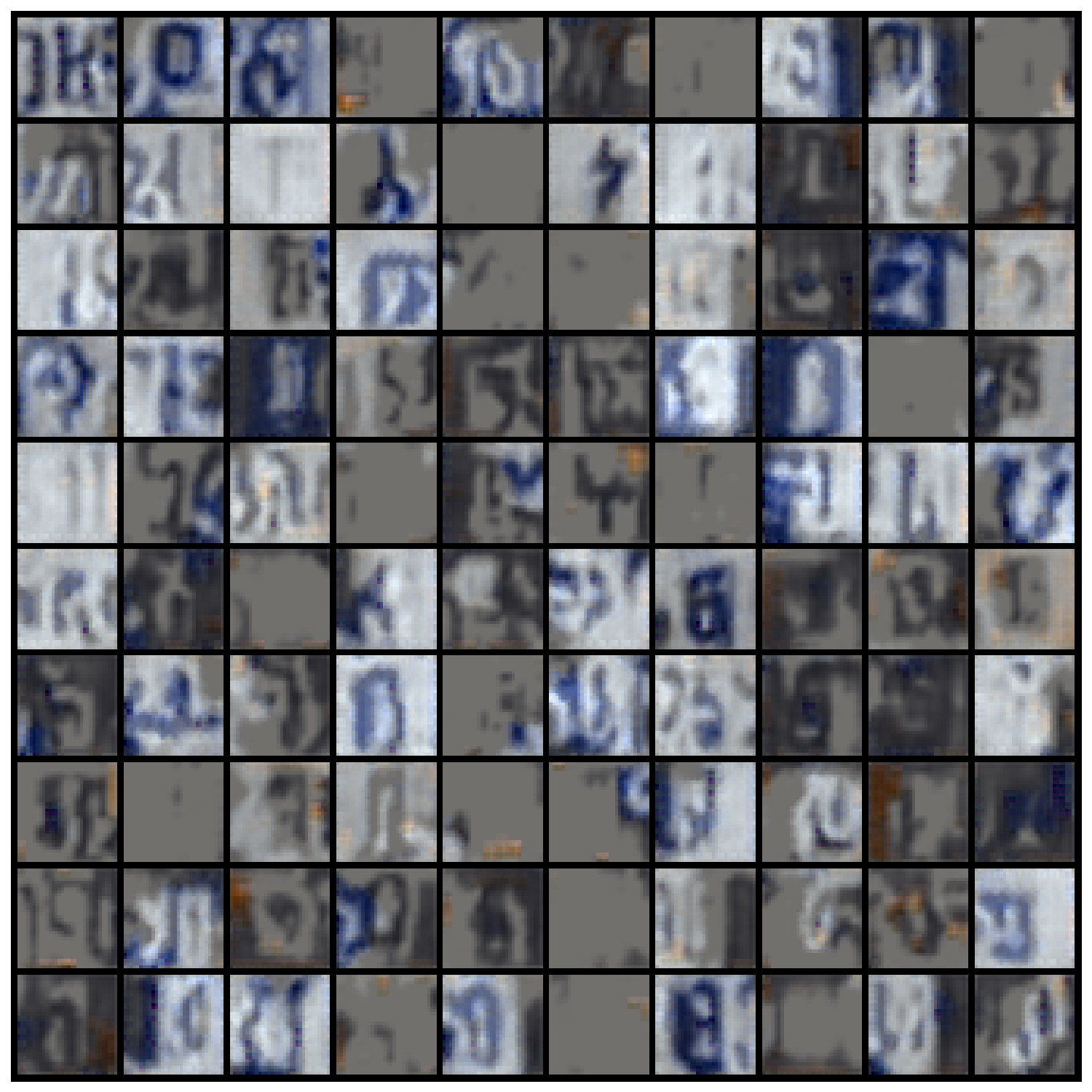}
    \centering Low AUM (Hard) data
  \end{minipage}\hfill
  \begin{minipage}{0.32\textwidth}
    \centering
    \includegraphics[width=1\textwidth]{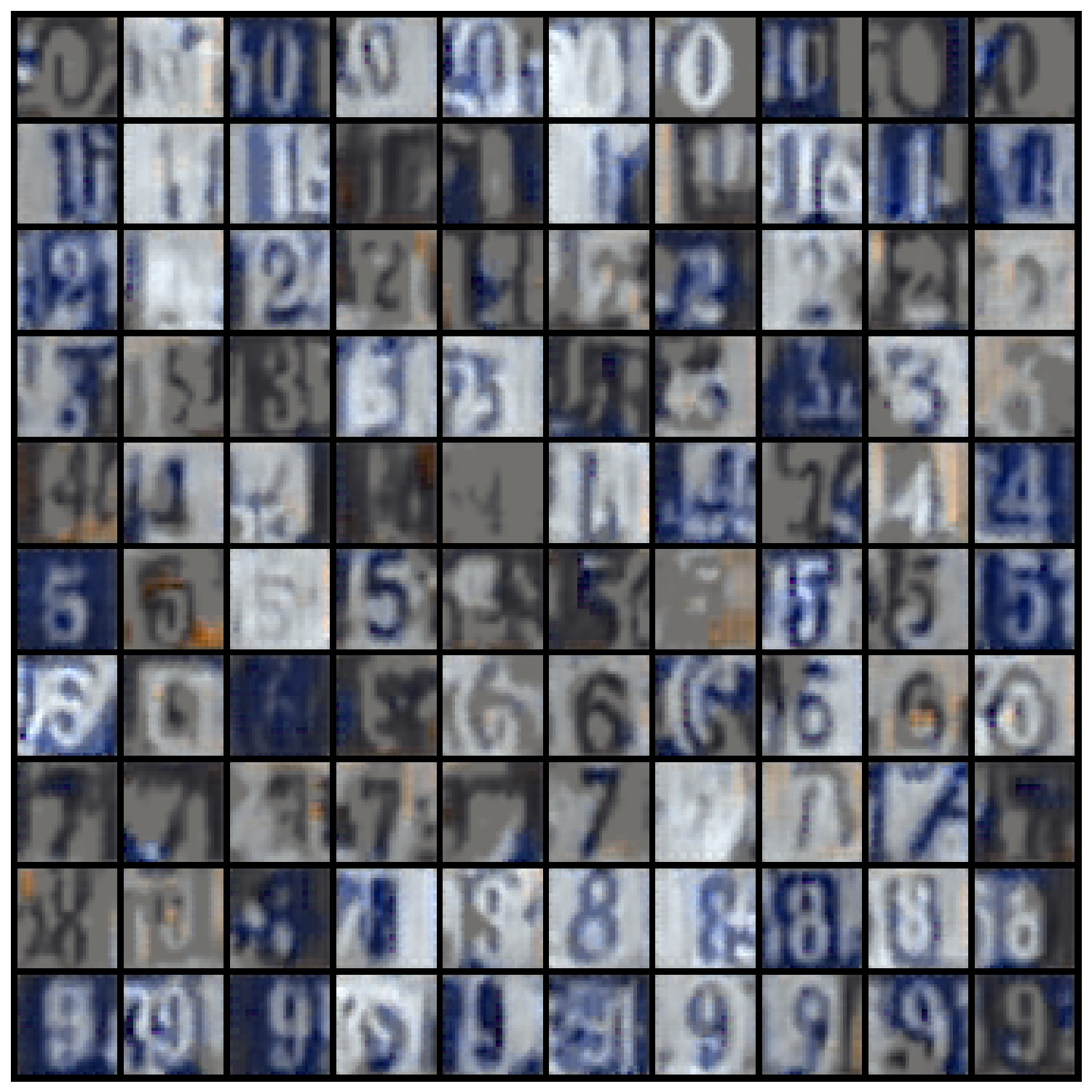}
    \centering Randomly selected data
  \end{minipage}
  \caption{Images generated by an SVHN HMN with 11IPC storage budget. Images in each row are from the same class.}
  \label{fig:svhn-10ipc}
\end{figure}

\begin{figure}[h]
  \centering
  \begin{minipage}{0.32\textwidth}
    \centering
    \includegraphics[width=1\textwidth]{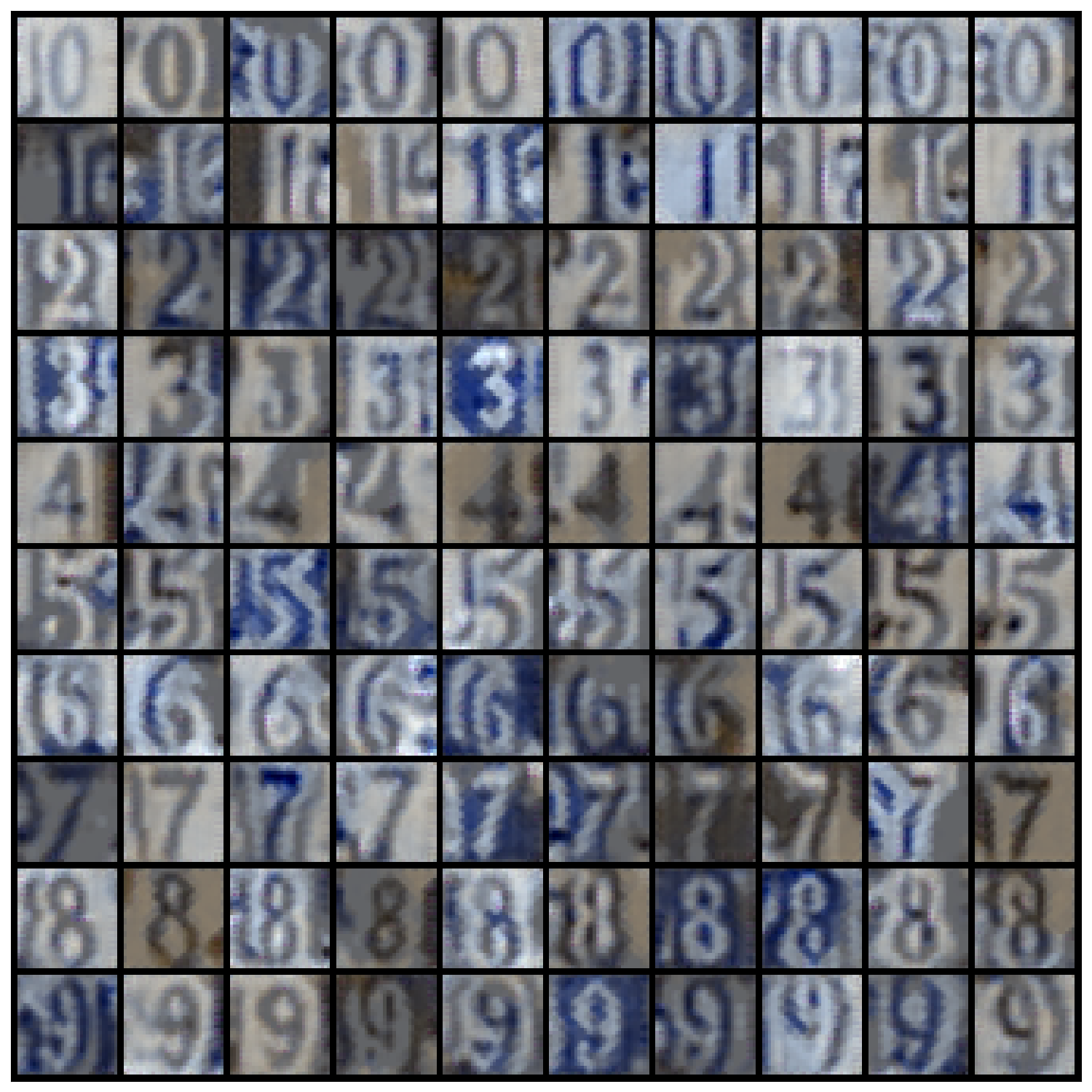}
    \centering High AUM (Easy) data
  \end{minipage}\hfill
  \begin{minipage}{0.32\textwidth}
    \centering
    \includegraphics[width=1\textwidth]{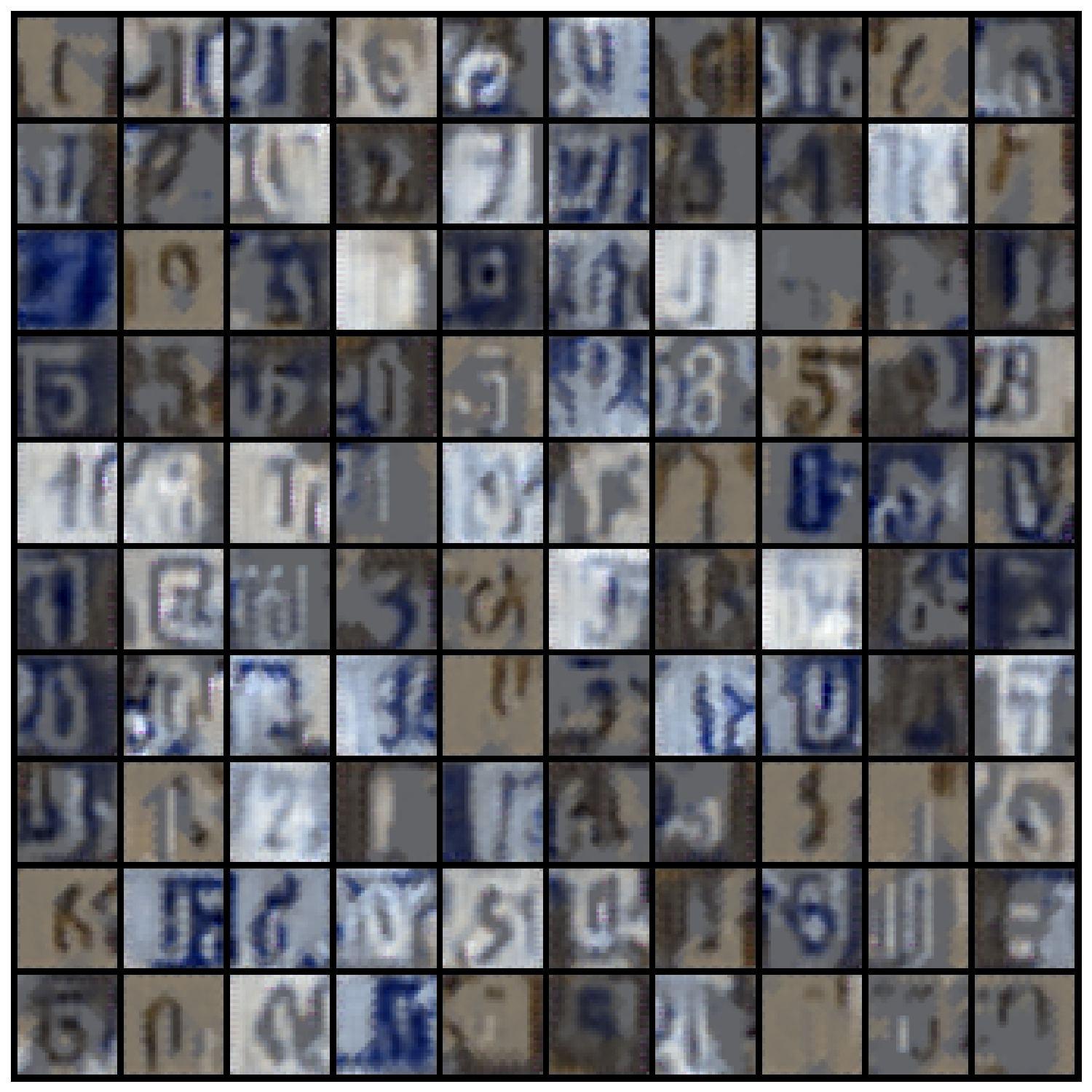}
    \centering Low AUM (Hard) data
  \end{minipage}\hfill
  \begin{minipage}{0.32\textwidth}
    \centering
    \includegraphics[width=1\textwidth]{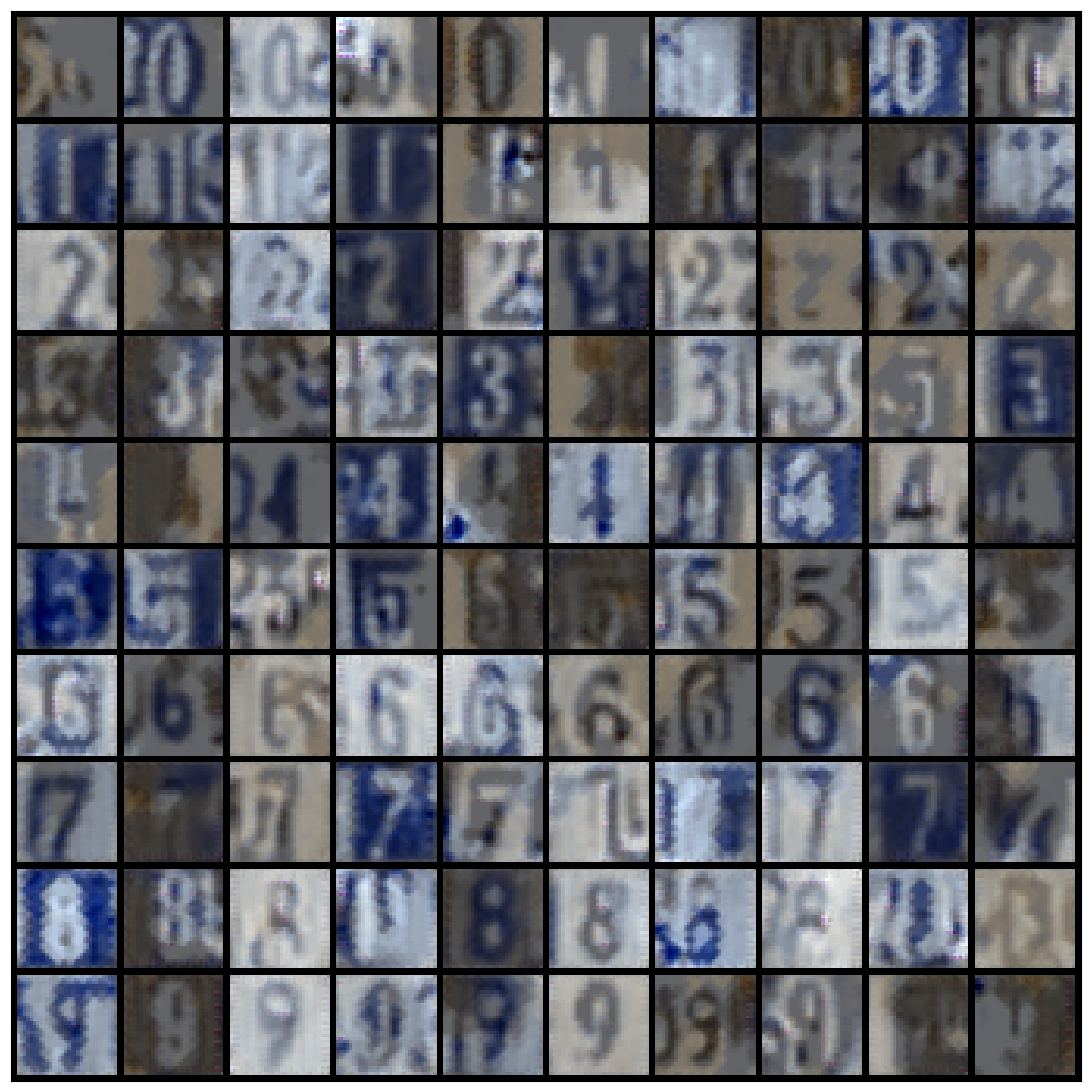}
    \centering Randomly selected data
  \end{minipage}
  \caption{Images generated by an SVHN HMN with 55IPC storage budget. Images in each row are from the same class.}
  \label{fig:svhn-50ipc}
\end{figure}

\end{document}